\pdfoutput=1

\documentclass[11pt]{article}

\usepackage[final]{acl}

\usepackage{times}
\usepackage{latexsym}

\usepackage[T1]{fontenc}
\usepackage{booktabs}
\usepackage[utf8]{inputenc}

\usepackage{microtype}

\usepackage{inconsolata}

\usepackage{graphicx}

\usepackage{kotex}
\usepackage{amsmath}
\usepackage{amssymb}
\usepackage{multirow}
\usepackage{adjustbox}
\usepackage{makecell}
\usepackage{float}
\usepackage[table]{xcolor}
\usepackage{array}
\usepackage{siunitx}
\usepackage{subcaption}
\usepackage{dblfloatfix}
\usepackage{placeins}
\usepackage{xurl}

\title{Detecting and Guiding LLM-Generated Korean Poetry\\with Interpretable Form-level Features}

\author{Keunhyeung Park$^{1}$, Seunguk Yu$^{1}$, YoungBin Kim$^{1,2}$ \\
  $^{1}$Department of Artificial Intelligence, Chung-Ang University \\
  $^{2}$Graduate School of Advanced Imaging Science, Multimedia \& Film, Chung-Ang University \\
  \texttt{synoark99@cau.ac.kr, seungukyu@gmail.com, ybkim85@cau.ac.kr}}

\begin{document}
\maketitle
\begin{abstract}
LLMs often struggle with modern Korean poetry, producing outputs that resemble ``line-broken prose.'' We address two coupled tasks: detecting whether a Korean poem is human- or LLM-authored, and guiding LLMs to generate poetry closer in form to human writing. We quantify the human--LLM gap along four form-level linguistic dimensions: output length (\textbf{\textit{Volume}}), the diversity and connective use of line-final forms (\textbf{\textit{Structure Variation}}), the irregularity of line lengths (\textbf{\textit{Rhythmic Irregularity}}), and adherence to standard orthography (\textbf{\textit{Normative Adherence}}). We operationalize these dimensions as five interpretable features. For detection, a logistic regression classifier over these five features attains an average AUC-ROC of 83.60 in zero-shot out-of-distribution detection across seven unseen LLMs, versus 75.84 for the strongest baseline in our comparison, KatFishNet, an absolute gain of 7.76 AUC points and a 10.23\% relative improvement; one generator-specific punctuation pattern outside our taxonomy remains a boundary case. For generation, expert evaluation on GPT-5.2 prefers feature-guided poems over the unconstrained baseline, and analyses across GPT-5.2 and Gemini-3 show that targeted length, rhythm, and ending statistics move toward the human distribution. These results suggest that interpretable, language-specific features can bridge the diagnosis and guidance of LLM-generated poetry\footnote{Our code and data are available at \url{https://github.com/keunhyeung/korean-poetry}.}.
\end{abstract}

\section{Introduction}
\label{sec:intro}
While large language models (LLMs) have demonstrated remarkable performance, they have exhibited limitations in poetry writing~\citep{hu2024poetrydiffusion}. However, their outputs are often described as ``line-broken prose'', favoring safe and plain word choices over poetic tension or rhythm~\citep{tian-peng-2022-zero}. This tendency is consistent with three documented sources of LLM bias: prose-dominant training corpora~\citep{brown2020languagemodelsfewshotlearners,touvron2023llamaopenefficientfoundation}, likelihood-maximization-induced diversity loss~\citep{NEURIPS2018_23ce1851}, and regression toward the mean~\citep{Shumailov2024}. In this regard, we characterize the form-level surface statistics associated with them in Korean poetry.

\begin{figure*}[t]
    \centering
    \includegraphics[width=0.95\textwidth]{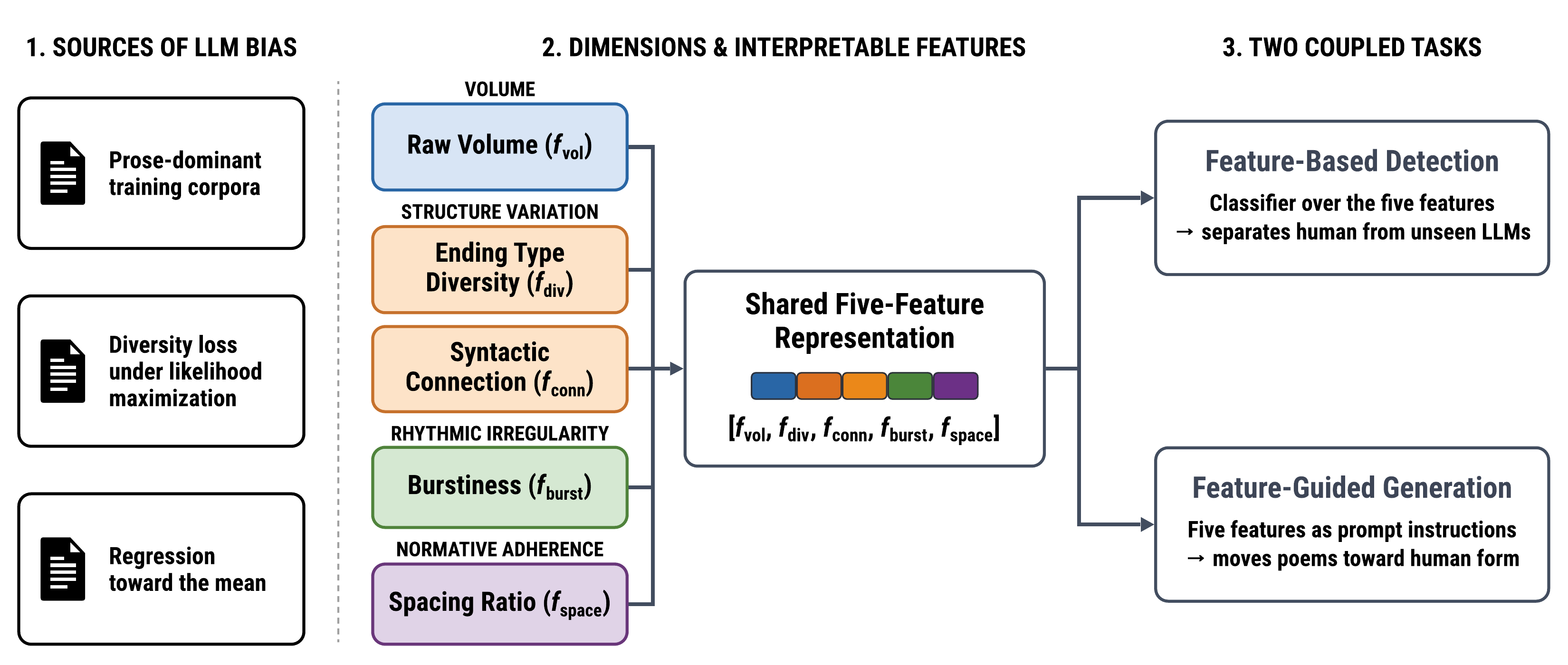}
    \caption{
    Overview of the introduced framework. Documented LLM biases motivate four form-level dimensions: \textbf{\textit{Volume}}, \textbf{\textit{Structure Variation}}, \textbf{\textit{Rhythmic Irregularity}}, and \textbf{\textit{Normative Adherence}}. We operationalize these as five interpretable features that support both zero-shot OOD detection and feature-guided generation.
    }
    \label{fig:overview}
\end{figure*}

Prior work on computational poetry generation has mainly relied on explicit constraints such as meter and rhyme~\citep{ghazvininejad-etal-2017-hafez,lau-etal-2018-deep,yu-etal-2024-charpoet}. However, such rigid constraints align poorly with modern Korean poetry, which often departs from fixed meter and builds rhythm through lineation and meaning-driven cadence~\citep{ART001269697,ART001234444,ART002200536}. We therefore take a diagnostic rather than constraint-engineering stance. Instead of designing a single mechanism to push outputs toward a desired form, we decompose the gap between human and LLM poetry into complementary axes for diagnosis and prompt-level intervention.

Drawing on prior work on LLM statistical uniformity and formal rigidity~\citep{Holtzman2020CuriousCase,doi:10.1073/pnas.2422455122,park-etal-2025-katfishnet}, we identify four linguistic dimensions that characterize form-level differences between human-authored and LLM-generated Korean poetry: (1) \textbf{\textit{Volume}}, (2) \textbf{\textit{Structure Variation}}, (3) \textbf{\textit{Rhythmic Irregularity}}, and (4) \textbf{\textit{Normative Adherence}}. We treat them as a diagnostic taxonomy motivated by the documented length, ending-form, rhythmic, and orthographic biases described above, and operationalize each dimension as a quantitative feature that doubles as a detection signal and an intervention target, as shown in Figure~\ref{fig:overview}.

Although our analysis focuses on Korean poetry, the broader method may inform work in other languages and genres by using interpretable language-specific features for both diagnosis and intervention. Our contributions are as follows:
\begin{itemize}
    \item We introduce a diagnostic taxonomy of four form-level dimensions for Korean LLM-generated poetry and operationalize them as five interpretable linguistic features.
    \item We show that these features support zero-shot out-of-distribution (OOD) detection across seven unseen LLMs, outperforming the strongest KatFishNet configuration on average while revealing generator-specific boundary cases. KatFishNet is the closest prior Korean LLM-text detector, uses punctuation and spacing cues, and provides our base dataset.
    \item We reuse the same features as prompt-level guidance and show that they improve expert judgments and shift several form-level statistics toward human poetry, while leaving orthographic regularity as a residual limitation.
\end{itemize}

\section{Related Work}
\label{sec:related_work}

\subsection[Linguistic Differences Between Human and LLM Text]{Linguistic Differences Between\protect\linebreak\hspace*{\parindent}Human and LLM Text}

Feature-based analyses indicate that LLM outputs occupy a constrained statistical space, with reduced variance in lexical diversity and sentence length compared to human writing~\citep{Opara24, Mu_oz_Ortiz_2024}. Building on these observations, recent studies show that generated texts exhibit stylistic uniformity through tight clustering and limited syntactic constructions, whereas human texts reflect heterogeneous individual variation~\citep{OSullivan2025, zamaraeva-etal-2025-comparing}.

Computational stylometry provides broad feature inventories for authorship, genre, readability, and text classification. Coh-Metrix~\citep{graesser2004cohmetrix} compiles cohesion, discourse, syntactic, lexical, and readability measures, while StyloMetrix~\citep{okulska2023stylometrix} operationalizes multilingual stylometric vectors. Recent stylometric detection work shows that such features can distinguish human and LLM-generated texts even in short samples~\citep{Przystalski_2026}. Poetry-analysis work further shows that formal cues such as line endings and verse structure are tied to particular languages and poetic traditions~\citep{desisto2024poetrysurvey, bhyravajjula-etal-2025-much}.

Human--LLM form contrasts are also visible in Korean LLM-generated text, where outputs default to explicit, low-context patterns~\citep{lee-etal-2025-testset}. The closest prior work to ours is KatFishNet~\citep{park-etal-2025-katfishnet}, which identifies rigid orthographic adherence through punctuation- and spacing-oriented cues and uses them as detection signals across Korean LLM text in general. Our work is complementary to these resources rather than a direct substitute, since we build a deliberately narrow, poetry-specific taxonomy for Korean free-verse lineation, endings, connectives, and spacing, and reuse it for both detection and generation guidance.

\subsection[Poetry Generation with Explicit Formal Constraints]{Poetry Generation with \protect\linebreak\hspace*{\parindent}Explicit Formal Constraints}

Prior work on computational poetry generation has primarily focused on explicit, hard constraints such as meter and rhyme. Early approaches enforced strict adherence by integrating formal constraints directly into the decoding process or optimization objectives~\citep{ghazvininejad-etal-2017-hafez, lau-etal-2018-deep}. To align text with planned structures under such rigid constraints, later work adopted hierarchical planning and dedicated constraints for acrostic poetry~\citep{tian-peng-2022-zero, agarwal-kann-2020-acrostic}. More recent studies use token-free or character-level modeling to enforce surface constraints such as exact character counts and rhyming schemes~\citep{belouadi-eger-2023-bygpt5, yu-etal-2024-charpoet}. Other approaches use diverse mechanisms, including unsupervised control codes and diffusion-based architectures, to manipulate metrical structures~\citep{ormazabal-etal-2022-poelm, hu2024poetrydiffusion}.

These methods are effective when the target form is defined by explicit templates, but they are less applicable to modern Korean poetry, which typically constructs rhythm through lineation and meaning-driven cadence rather than fixed meters~\citep{ART001269697, ART001234444}. Other lines of work attempt to internalize such formal constraints through parameter-efficient fine-tuning~\citep{hu2022lora}, including Czech poetry with metre and rhyme annotations~\citep{rosa-etal-2025-edupo} and Sanskrit Anushtubh-meter generation~\citep{jagadeeshan-etal-2026-chandomitra}. Rather than enforcing rigid formal constraints or learning them through weight updates, we guide generation using soft linguistic signals derived from cues that distinguish human poems from LLM outputs.

\section[Linguistic Dimensions: Four Key Differences]{Linguistic Dimensions: \protect\linebreak\hspace*{\parindent}Four Key Differences}
\label{sec:fingerprints}
We identify four linguistic dimensions that differentiate human-authored Korean poetry from LLM-generated poetry: \textbf{\textit{Volume}}, \textbf{\textit{Structure Variation}}, \textbf{\textit{Rhythmic Irregularity}}, and \textbf{\textit{Normative Adherence}}. We treat them as a diagnostic decomposition motivated by regularities documented in prior work, not as causal mechanisms established here.

Length inflation from prose-dominant pre-training motivates \textbf{\textit{Volume}}. Diversity loss under likelihood maximization motivates \textbf{\textit{Structure Variation}} through convergence on a small set of sentence-final and connective forms. Regression toward the mean in line length motivates \textbf{\textit{Rhythmic Irregularity}}. Orthographic standards reinforced by the same prose-dominant corpora motivate \textbf{\textit{Normative Adherence}}. We operationalize these dimensions with the five quantitative features below. Appendix~\ref{sec:appendix_notation} defines the variables.

\subsection{Volume}
\noindent \textbf{Raw Volume ($f_{\text{vol}}$)} Humans and LLMs differ significantly in their preferred output length~\citep{ijcai2025p1144}. We capture this difference with a token-count measure. Let $x$ denote the input poem and $\tau(\cdot)$ represent the whitespace tokenization function. Raw volume is defined as the cardinality of the tokenized sequence:
\begin{equation}
    f_{\text{vol}} = |\tau(x)|,
    \label{eq:volume}
\end{equation}

\subsection{Structure Variation}

\noindent \textbf{Ending Type Diversity ($f_{\text{div}}$)} Human-authored poems tend to employ a wider range of sentence-final forms even within limited length, whereas LLM outputs often repeat a limited set of endings~\citep{Holtzman2020CuriousCase,zamaraeva-etal-2025-comparing}. We extract the ending sequence $\mathcal{E}$ from all line-level units in a poem. For each line, the Kkma part-of-speech tagger~\citep{park2014konlpy} identifies the final token. Each ending is represented as a normalized line-final form combining its surface suffix and POS category rather than only the coarse \textit{final}/\textit{connective}/\textit{other} label. The coarse label is reserved for $f_{\text{conn}}$. If no explicit line breaks exist, we parse sentence-like units delimited by punctuation or newlines. Diversity is the ratio of distinct ending types to the total count:
\begin{equation}
    f_{\text{div}} = \frac{|\mathcal{U}_E|}{|\mathcal{E}|},
    \label{eq:ending_diversity}
\end{equation}
Here, $|\mathcal{E}|$ counts line-final endings, one per line-level unit, and $|\mathcal{U}_E|$ counts their distinct types.

\noindent \textbf{Syntactic Connection ($f_{\text{conn}}$)} Korean authors often allow implicit relations between clauses and lines~\citep{hall1976beyond,Lincoln2010FarSide,park-etal-2015-zero}, whereas LLM outputs more often make discourse relations explicit~\citep{lee-etal-2025-testset}. Using the same unit-level ending sequence $\mathcal{E}$, we classify an ending as \textit{connective} if its POS tag is a connective-ending tag or if it matches a heuristic set of common Korean connective suffixes. Syntactic connection is then quantified as the ratio of connective endings:
\begin{equation}
    f_{\text{conn}} = \frac{1}{|\mathcal{E}|} \sum_{e \in \mathcal{E}} \mathbb{I}(e = \text{conn}),
    \label{eq:syn_conn}
\end{equation}

\begin{table*}[t!]
\centering
\setlength{\tabcolsep}{3pt}
\renewcommand{\arraystretch}{1.15}

\resizebox{0.95\textwidth}{!}{%
\begin{tabular}{lcccccccc}
\toprule
\textbf{Detection Methods} &
\textbf{Qwen2-72B} &
\textbf{Solar} &
\textbf{Llama-3.1-70B} &
\textbf{Gemini-3} &
\textbf{GPT-5.2} &
\textbf{\makecell{EXAONE-3.5-7.8B}} &
\textbf{EEVE-10.8B} &
\textbf{Average} \\
\midrule
KatFishNet (\makecell[l]{Punctuation}) & 93.45 & 62.65 & 63.22 & 59.48 & 58.76 & 81.38 & 75.10 & 70.58 \\
KatFishNet (\makecell[l]{Punctuation\\+ Spacing}) & \textbf{95.41} & 73.78 & 58.64 & 64.88 & 71.31 & 87.12 & \textbf{79.77} & 75.84 \\
\textsc{\textbf{Ours}} & 79.78 & \textbf{86.39} & \textbf{73.22} & \textbf{82.73} & \textbf{96.42} & \textbf{89.85} & 76.84 & \textbf{83.60} \\
\bottomrule
\end{tabular}%
}

\caption{Zero-shot OOD AUC-ROC across target LLMs. Our five features average 83.60, compared with 75.84 for the strongest KatFishNet configuration. Bold marks the best result for each model.}
\label{tab:main_logistic}
\end{table*}

\begin{table}[t!]
    \centering
    \small
    \begin{tabular}{lcc}
        \toprule
        \textbf{Source} & \textbf{Size} & \textbf{Origin} \\
        \midrule
        Human & 189 & KatFishNet \\
        GPT-4o & 189 & KatFishNet \\
        Solar & 189 & KatFishNet \\
        Qwen2-72B & 189 & KatFishNet \\
        Llama-3.1-70B & 189 & KatFishNet \\
        \midrule
        Gemini-3 & 189 & Ours \\
        GPT-5.2 & 189 & Ours \\
        EXAONE-3.5-7.8B & 181 & Ours \\
        EEVE-10.8B & 177 & Ours \\
        \bottomrule
    \end{tabular}
    \caption{Dataset composition. ``KatFishNet'' denotes the original benchmark data, while ``Ours'' denotes newly collected and screened poems.}
    \label{tab:dataset_stats}
\end{table}

\subsection{Rhythmic Irregularity}
\noindent \textbf{Burstiness ($f_{\text{burst}}$)} Modern Korean poetry often departs from fixed meters and instead develops rhythm through phrasing and breath~\citep{ART001269697,ART001234444}, whereas LLM outputs tend to produce more uniform sentence structures and line-length patterns~\citep{yadagiri-etal-2024-detecting, Mu_oz_Ortiz_2024}. More broadly, LLM-generated text can exhibit reduced dispersion and limited variability compared to human writing~\citep{Opara24,zamaraeva-etal-2025-comparing}. To quantify rhythmic variability, we construct the length array $\mathcal{L}$, where each element is the number of content characters in a line, excluding whitespace and punctuation, and compute the coefficient of variation (CV):
\begin{equation}
    f_{\text{burst}} = \frac{\sigma_{\mathcal{L}}}{\mu_{\mathcal{L}}},
    \label{eq:burstiness}
\end{equation}
LLMs typically generate uniform, low-CV line lengths, mimicking rigid prose or fixed meters, whereas human poets actively manipulate line breaks based on phrasing and breath, yielding a high-CV mix of short and long lines. $\text{CV}(\mathcal{L})$ thus serves as a quantifiable proxy for breath-driven structural irregularity.

\subsection{Normative Adherence}
\noindent \textbf{Spacing Ratio ($f_{\text{space}}$)} LLM outputs tend to follow standard orthography and spacing conventions more consistently, whereas human authors may deviate from standard spacing intentionally or unintentionally due to poetic practice~\citep{park-etal-2025-katfishnet,wu-etal-2025-wrote}. We measure normative rigidity as the ratio of space characters to total characters:
\begin{equation}
    f_{\text{space}} = \frac{N_{\text{space}}}{N_{\text{char}}},
    \label{eq:spacing}
\end{equation}

\section{Detection via Linguistic Features}
\label{sec:exp1}
\begin{table}[t!]
\centering
\setlength{\tabcolsep}{4pt}
\renewcommand{\arraystretch}{1.15}
\resizebox{\columnwidth}{!}{%
\begin{tabular}{lccccl}
\toprule
\textbf{Target LLM} & \textbf{KatFishNet} & \textbf{\textsc{Ours}} & \textbf{$\Delta$(\%)} & $p_{\text{holm}}$ & \textbf{Result} \\
\midrule
Qwen2-72B & 95.41 & 79.78 & -16.38 & $<$0.001 & Base \\
Solar & 73.78 & 86.39 & +17.09 & $<$0.001 & \textsc{Ours} \\
Llama-3.1-70B & 58.64 & 73.22 & +24.87 & $<$0.001 & \textsc{Ours} \\
Gemini-3 & 64.88 & 82.73 & +27.51 & $<$0.001 & \textsc{Ours} \\
GPT-5.2 & 71.31 & 96.42 & +35.21 & $<$0.001 & \textsc{Ours} \\
EXAONE-3.5-7.8B & 87.12 & 89.85 & +3.13 & 0.0799 & Tie \\
EEVE-10.8B & 79.77 & 76.84 & -3.67 & 0.8041 & Tie \\
\bottomrule
\end{tabular}%
}
\caption{Paired DeLong tests comparing our method with KatFishNet using punctuation and spacing. Holm correction is applied, and our gain is significant on four of seven targets.}
\label{tab:delong}
\end{table}

\begin{table*}[!tbp]
\centering
\small
\setlength{\tabcolsep}{3pt}
\renewcommand{\arraystretch}{1.15}

\resizebox{0.95\textwidth}{!}{%
\begin{tabular}{lcccccccc}
\toprule
\textbf{Detection Methods} &
\textbf{Qwen2-72B} &
\textbf{Solar} &
\textbf{Llama-3.1-70B} &
\textbf{Gemini-3} &
\textbf{GPT-5.2} &
\textbf{\makecell{EXAONE-3.5-7.8B}} &
\textbf{EEVE-10.8B} &
\textbf{Average} \\
\midrule
\textsc{Ours} & 79.78 & 86.39 & \textbf{73.22} & \textbf{82.73} & 96.42 & \textbf{89.85} & 76.84 & \textbf{83.60} \\
\makecell[l]{\textminus\ Volume} & \textbf{80.48} & 85.87 & 68.55 & 80.62 & 95.58 & 89.32 & \textbf{81.10} & 83.07 \\
\makecell[l]{\textminus\ Rhythmic Irregularity} & 78.73 & 77.87 & 71.02 & 81.91 & \textbf{97.15} & 84.21 & 69.56 & 80.06 \\
\makecell[l]{\textminus\ Normative Adherence} & 75.19 & \textbf{87.01} & 65.18 & 76.04 & 74.87 & 82.70 & 77.22 & 76.89 \\
\makecell[l]{\textminus\ Structure Variation} & 67.22 & 74.12 & 67.64 & 65.55 & 78.40 & 80.54 & 69.56 & 71.86 \\
\bottomrule
\end{tabular}%
}

\caption{AUC-ROC after removing each linguistic dimension. Every removal lowers the mean score. Bold marks the best result for each model.}
\label{tab:main_ablation}
\end{table*}

\begin{table*}[t!]
  \centering
  \begin{minipage}[t]{0.46\textwidth}
  \centering
  \small
  \setlength{\tabcolsep}{5pt}
  \renewcommand{\arraystretch}{1.40}
  \resizebox{\linewidth}{!}{%
  \begin{tabular}{l c r r r}
  \toprule
  \textbf{Feature} & \textbf{Sign} & \textbf{Avg} & \textbf{Min} & \textbf{Max} \\
  \midrule
  Ending Type Diversity & $+$ & 1.056 & 0.940 & 1.150 \\
  Burstiness            & $+$ & 1.003 & 0.896 & 1.078 \\
  Spacing Ratio         & $-$ & -0.778 & -0.869 & -0.659 \\
  Raw Volume            & $+$ & 0.559 & 0.450 & 0.675 \\
  Syntactic Connection  & $-$ & -0.494 & -0.542 & -0.463 \\
  \bottomrule
  \end{tabular}%
  }
  \captionof{table}{Logistic Regression coefficient statistics across per-model classifiers. Sign denotes coefficient direction, which is consistent across all seven target models. Avg/Min/Max are standardized coefficients.}
  \label{tab:main_weight}
  \end{minipage}\hspace{0.03\textwidth}%
  \begin{minipage}[t]{0.46\textwidth}
  \centering
  \small
  \setlength{\tabcolsep}{5pt}
  \renewcommand{\arraystretch}{1.15}
  \resizebox{\linewidth}{!}{%
  \begin{tabular}{lcc}
  \toprule
  \textbf{Target LLM} & \textbf{MMD (Features)} & \textbf{MMD (Embeddings)} \\
  \midrule
  Qwen2-72B & 0.3791 & 0.2967 \\
  Solar & 0.3958 & 0.2342 \\
  Llama-3.1-70B & 0.2423 & 0.1773 \\
  Gemini-3 & 0.3695 & 0.2000 \\
  GPT-5.2 & 0.5919 & 0.2911 \\
  EXAONE-3.5-7.8B & 0.4321 & 0.3101 \\
  EEVE-10.8B & 0.3479 & 0.2521 \\
  \midrule
  \textbf{Average} & 0.3941 & 0.2516 \\
  \bottomrule
  \end{tabular}%
  }
  \captionof{table}{Maximum Mean Discrepancy (MMD) for our five features and embeddings. Raw magnitudes are not compared because the representations differ in dimensionality and scale.}
  \label{tab:mmd}
  \end{minipage}
\end{table*}

\subsection{Experimental Setup}

We build on the KatFishNet dataset~\citep{park-etal-2025-katfishnet}, which comprises 945 poems with 189 poems per source from human writers and four base LLMs (GPT-4o, Solar, Qwen2-72B, and Llama-3.1-70B). To broaden coverage, we add comparably sized samples from four additional models (Gemini-3, GPT-5.2, EXAONE-3.5-7.8B, and EEVE-10.8B). All newly generated poems underwent manual screening to exclude outputs that violated prompting instructions, exhibited degraded quality, or contained meta-commentary, as summarized in Table~\ref{tab:dataset_stats}. After screening, the final counts are 189, 189, 181, and 177 respectively. Further details including per-model screening are provided in Appendix~\ref{app:md_details}.

KatFishNet is the closest prior work in this setting. We follow its OOD protocol for direct comparability, but evaluate a different feature philosophy and analyze where KatFishNet-style cues outperform ours. The OOD protocol measures generalization to generators not seen during training. The backbone classifier is logistic regression operating on the five-dimensional feature vector derived from our linguistic dimensions. Following KatFishNet, we split human-authored poems into 8:2 for training and evaluation. Training combines the 80\% human subset with GPT-4o-generated poems, and each of the seven unseen LLMs is evaluated on a separate test set that pairs its poems with the held-out 20\% human poems.

\subsection{Zero-Shot OOD Detection}

Table~\ref{tab:main_logistic} reports the zero-shot OOD detection results. Our approach reaches an average AUC-ROC of 83.60 across the seven evaluation models, a 10.23\% relative improvement over the best-performing KatFishNet configuration, which averages 75.84 under the same protocol. On GPT-5.2 and Gemini-3, the method achieves 96.42 and 82.73, relative gains of 35.21\% and 27.51\% over KatFishNet.

Table~\ref{tab:delong} reports paired DeLong tests~\citep{delongtest} using Stouffer's method across five seeds with Holm correction over seven targets. Gains are significant for Solar, Llama-3.1-70B, Gemini-3, and GPT-5.2. Results are not significant for EXAONE-3.5-7.8B or EEVE-10.8B. KatFishNet performs better on Qwen2-72B, where it captures a generator-specific comma pattern outside our length, structure, rhythm, and spacing features. Appendix~\ref{sec:appendix_qwen2} analyzes this boundary case, and Appendix~\ref{sec:appendix_fusion} reports feature fusion.

\subsection{Feature Validity}

Table~\ref{tab:main_ablation} shows that ablating any of the four dimension groups reduces average AUC-ROC, with Structure Variation removal causing the largest drop. No single group in isolation matches the full system, and dimension-wise breakdowns are provided in Appendix~\ref{sec:appendix_concept_wise}.

Table~\ref{tab:main_weight} reports standardized coefficients from a separate logistic regression classifier per target LLM, where every feature's coefficient sign remains constant across generators. Following the interpretability of linear coefficients~\citep{Molnar_2020} and feature-stability criteria in authorship attribution~\citep{stamatatos2009survey}, this sign consistency indicates that the signals are stable across generator change rather than artifacts of a particular model.

Table~\ref{tab:mmd} reports two-sample Maximum Mean Discrepancy (MMD) tests for our five features and \texttt{klue/roberta-large} embeddings~\citep{park2021klue}, using a radial basis function (RBF) kernel~\citep{gretton2012kernel}. MMD magnitudes are not comparable across representations, so we test each representation separately. Within each representation, the observed MMD for every target exceeds all values from 1{,}000 label shuffles and remains significant after Holm correction over the seven targets. This confirms a distributional gap between human and LLM poems in both representations.


\subsection{Robustness}
\label{sec:robustness}

We use two robustness checks to test whether the feature representation depends on a particular training generator or decoding setting. Table~\ref{tab:cross_source_summary} shows that, across each of the eight available LLMs as the training source, our method averages 80.44 AUC-ROC, compared with 70.34 for KatFishNet. Appendix~\ref{sec:appendix_cross_source} provides the full cross-source matrices.

Across GPT-5.2 decoding temperatures from 0.5 to 1.2, the GPT-4o-trained classifier remains above 95 AUC-ROC without retraining, ranging from 95.63 to 98.12 in Appendix~\ref{sec:appendix_temperature}. Together with the cross-source result, this indicates stability across training generators and decoding settings.

Feature-level contrasts between human and LLM poetry are further reported in Appendix~\ref{sec:appendix_feature_analysis}.

\begin{table}[t]
\centering
\small
\setlength{\tabcolsep}{5pt}
\renewcommand{\arraystretch}{1.15}
\begin{tabular}{lcc}
\toprule
\textbf{Train Source} & \textbf{\textsc{Ours}} & \textbf{KatFishNet} \\
\midrule
Qwen2-72B & 81.96 & 68.90 \\
Solar & 79.53 & 72.66 \\
Llama-3.1-70B & 75.77 & 72.50 \\
Gemini-3 & 83.67 & 70.73 \\
GPT-5.2 & 73.73 & 69.35 \\
EXAONE-3.5-7.8B & 83.13 & 66.16 \\
EEVE-10.8B & 82.10 & 66.58 \\
GPT-4o & 83.60 & 75.84 \\
\midrule
\textbf{Average} & \textbf{80.44} & 70.34 \\
\bottomrule
\end{tabular}
\caption{Average cross-source AUC-ROC by training source. Each row trains on one LLM and evaluates on all remaining targets.}
\label{tab:cross_source_summary}
\end{table}

\section{Feature-Guided Generation}
\label{sec:exp2}

Feature-guided prompting improves expert judgments of poetic naturalness and shifts form-level statistics toward the human distribution. Reduced detector separability serves only as supporting evidence. Following the property-centric framework of \citet{long-etal-2025-makes}, we translate the five diagnostic features into prompt instructions. We compare an unconstrained \textsc{Baseline} with a feature-guided \textsc{Refined} condition. Automatic analyses cover GPT-5.2 and Gemini-3, while the blind expert ranking covers GPT-5.2.

\subsection{Feature-Guided Generation Framework}
\label{sec:promptframwork}

\textbf{Generation Conditions.} We use the poetry generation framework of \citet{park-etal-2025-katfishnet} as our \textsc{Baseline}. It conditions on a human-authored reference poem and target age group without imposing low-level linguistic constraints. \textsc{Refined} keeps this content conditioning fixed and adds guidance derived from our four linguistic dimensions.

\textbf{Prompt Guidance.} We translate the four linguistic dimensions into five concrete instructions. \textit{Volume Synchronization} asks the model to match the reference poem's length. \textit{Rhythmic Variation} permits non-uniform line lengths. \textit{Connective Reduction} discourages frequent Korean connective endings in favor of line breaks. \textit{Ending Diversity} asks for varied sentence-final forms, including nominal endings. \textit{Flexible Spacing} permits minor deviations from standard orthography. Appendix~\ref{sec:prompt} provides the full prompt, and Appendix~\ref{sec:appendix_prompt_robustness} reports prompt sensitivity and per-instruction ablation.

\begin{table}[t!]
\centering
\small
\setlength{\tabcolsep}{3.5pt}
\renewcommand{\arraystretch}{1.2}

\begin{tabular}{lcccccc}
\toprule
\textbf{Category} & \textbf{E1} & \textbf{E2} & \textbf{E3} & \textbf{E4} & \textbf{E5} & \textbf{Avg} \\
\midrule
\textsc{Baseline} & 31 & 32 & 16 & 26 & 26 & 26.2 \\
\textsc{Refined}  & \textbf{46} & 42 & 49 & 45 & 43 & 45.0 \\
\textsc{Human}    & 43 & \textbf{46} & \textbf{55} & \textbf{49} & \textbf{51} & \textbf{48.8} \\
\bottomrule
\end{tabular}

\caption{Aggregated Borda scores from five evaluators over 40 triplets. \textsc{Baseline} and \textsc{Refined} are GPT-5.2 outputs. First-, second-, and third-ranked poems receive 2, 1, and 0 points, with a maximum of 80 per evaluator. \textsc{Refined} ranks above \textsc{Baseline} and closer to \textsc{Human}. \textbf{E} denotes evaluator.}
\label{tab:main_human}
\end{table}

\subsection{Evaluation of Feature-Guided Generation}

\subsubsection[Expert Evaluation: Perceived Naturalness]{Expert Evaluation: \protect\linebreak\hspace*{\parindent}Perceived Naturalness}
\label{sec:main_expert_evaluation}
We conduct a blind evaluation with five native Korean speakers majoring in Korean literature. All generated poems use GPT-5.2 under the \textsc{Baseline} and \textsc{Refined} conditions. We sample 40 triplets containing \textsc{Human}, \textsc{Baseline}, and \textsc{Refined} poems and randomize their order within each triplet. Appendix~\ref{sec:human_eval_questionnaire} provides the questionnaire.

Evaluators rank the three poems by how naturally they read as human-authored poetry. We aggregate rankings using Borda Count~\citep{himmi-etal-2024-towards}, assigning 2 points to the first-ranked poem, 1 point to the second, and 0 points to the third.

Table~\ref{tab:main_human} reports pooled scores across 40 triplets. Each triplet is the paired unit, with the five evaluators' Borda scores summed by condition. Mean Borda totals per evaluator are 26.2 for \textsc{Baseline}, 45.0 for \textsc{Refined}, and 48.8 for \textsc{Human}. The Friedman test is significant at $p < 0.001$. Bonferroni-corrected Wilcoxon tests place both \textsc{Refined} and \textsc{Human} above \textsc{Baseline} at $p < 0.001$. \textsc{Refined} and \textsc{Human} do not differ significantly at $p = 1.0$, but this does not establish equivalence. We therefore claim only that guidance narrows the observed gap. Appendix~\ref{sec:appendix_friedman} reports the rank distributions. Inter-annotator agreement on the pooled rankings is Krippendorff's $\alpha = 0.378$, consistent with ranking-based creative-text evaluation~\citep{clark-etal-2021-thats, marco-etal-2025-reader}, and is discussed in Appendix~\ref{app:kripp}.

\subsubsection[Linguistic Shifts under Feature Guidance]{Linguistic Shifts under \protect\linebreak\hspace*{\parindent}Feature Guidance}
\label{sec:quant_drivers}

We report distributional statistics aggregated across Gemini-3 and GPT-5.2. Table~\ref{tab:main_refined_conn} shows that mean $f_{\text{conn}}$ falls from 0.357 under \textsc{Baseline} to 0.248 under \textsc{Refined}, moving toward the \textsc{Human} mean of 0.206.

\begin{figure}[t!]
    \centering
    \includegraphics[width=0.95\linewidth]{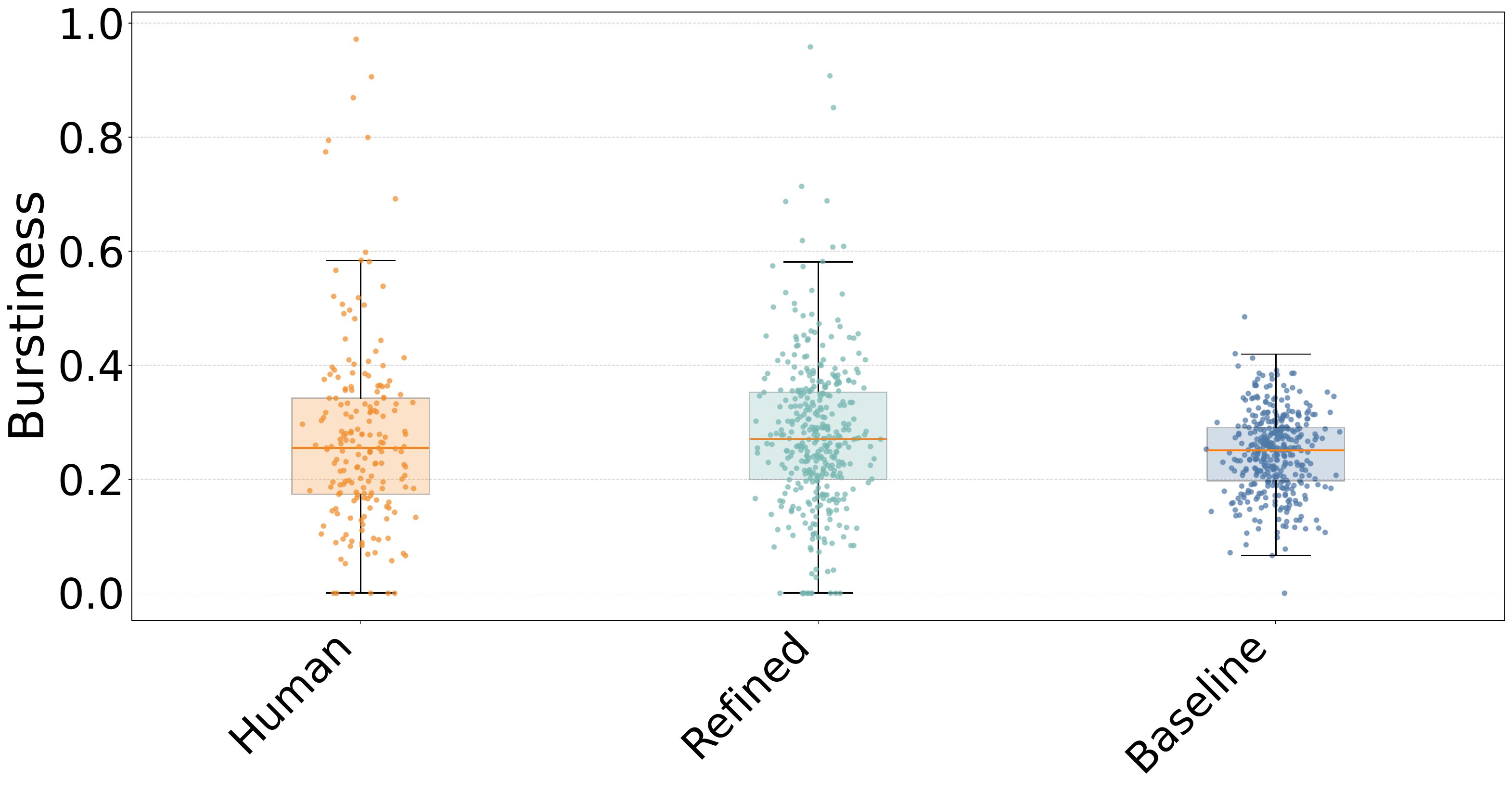}
    \caption{Distribution of $f_{\text{burst}}$ for human-authored poems and poems generated under \textsc{Baseline} and \textsc{Refined} aggregated across Gemini-3 and GPT-5.2.}
    \label{fig:burst_distribution}
\end{figure}

\begin{table}[t!]
    \centering
    \begin{tabular}{lcc}
    \toprule
    \textbf{Source} & \textbf{Mean} & \textbf{Std} \\
    \midrule
    \textsc{Baseline} & 0.357 & 0.154 \\
    \textsc{Refined} & 0.248 & 0.173 \\
    \midrule
    \textsc{Human} & 0.206 & 0.186 \\
    \bottomrule
    \end{tabular}
    \caption{Mean Syntactic Connection $f_{\text{conn}}$ across generation conditions. The generated conditions pool Gemini-3 and GPT-5.2 outputs.}
    \label{tab:main_refined_conn}
\end{table}


\definecolor{evasivegray}{gray}{0.94}
\definecolor{deltagray}{gray}{0.88}

\newcolumntype{M}{>{\centering\arraybackslash}p{4.2cm}}

\newcolumntype{E}{>{\columncolor{evasivegray}\centering\arraybackslash}c}

\newcolumntype{D}{>{\columncolor{deltagray}\centering\arraybackslash}c}

\begin{table*}[!tbp]
\centering
\small
\setlength{\tabcolsep}{7pt}
\renewcommand{\arraystretch}{1.34}

\begin{tabular}{M      c        E       D          c        E       D}
\toprule
\multirow{2}{*}{\textbf{Detection Method}} &
\multicolumn{3}{c}{\textbf{Gemini-3}} &
\multicolumn{3}{c}{\textbf{GPT-5.2}} \\
\cmidrule(lr){2-4}\cmidrule(lr){5-7}
& \textsc{\textbf{Baseline}} & \textsc{\textbf{Refined}} & \textbf{$\Delta$(\%)}
& \textsc{\textbf{Baseline}} & \textsc{\textbf{Refined}} & \textbf{$\Delta$(\%)} \\
\midrule

\makecell[c]{Log-Likelihood\\\citep{solaiman2019releasestrategiessocialimpacts}}
& 26.82 & 15.57 & -41.95 & 59.28 & 36.71 & -38.07 \\
\midrule

\makecell[c]{Rank\\\citep{gehrmann-etal-2019-gltr}}
& 13.85 & 18.47 & +33.36 & 47.32 & 32.06 & -32.25 \\
\midrule

\makecell[c]{Entropy\\\citep{gehrmann-etal-2019-gltr}}
& 18.16 & 12.25 & -32.54 & 43.28 & 31.29 & -27.70 \\
\midrule

\makecell[c]{LRR\\\citep{su-etal-2023-detectllm}}
& 19.44 & 16.07 & -17.34 & 34.86 & 26.96 & -22.66 \\
\midrule

\makecell[c]{NPR\\\citep{su-etal-2023-detectllm}}
& 47.31 & 37.79 & -20.12 & 81.70 & 71.16 & -12.90 \\
\midrule

\makecell[c]{Binoculars\\\citep{pmlr-v235-hans24a}}
& 59.67 & 39.67 & -33.52 & 67.49 & 43.17 & -36.03 \\
\midrule

\makecell[c]{Fast-DetectGPT\\\citep{bao2024fastdetectgpt}}
& 59.30 & 38.67 & -34.79 & 87.12 & 62.59 & -28.16 \\
\midrule

\textsc{\textbf{Ours}}
& 82.73 & 57.67 & -30.29 & 96.42 & 84.28 & -12.59 \\

\bottomrule
\end{tabular}

\caption{Detector AUC-ROC under \textsc{Baseline} and \textsc{Refined} prompts. Most detectors decrease under \textsc{Refined}. Raw AUC values follow each detector's score orientation, so values below 50 may reflect reversal rather than weak separability. LRR and NPR denote Log-Likelihood Log-Rank Ratio and Normalized Perturbed log-Rank.}
\label{tab:main_detection}
\end{table*}

Figure~\ref{fig:burst_distribution} shows a parallel shift in rhythm. Under \textsc{Refined}, the distribution of $f_{\text{burst}}$ moves toward \textsc{Human} and becomes less concentrated than \textsc{Baseline}. This shift is consistent with the expert naturalness rankings.

The shift varies by dimension. Volume, burstiness, ending diversity, and connectivity close between 72\% and 107\% of the gap from \textsc{Baseline} to \textsc{Human}, with the connective ratio at the lower end as derived from Table~\ref{tab:main_refined_conn}. Appendix~\ref{refined_app:feature_analysis} reports the remaining feature shifts. By contrast, spacing closes only 13\%, as its mean changes from 0.243 under \textsc{Baseline} to 0.236 under \textsc{Refined} while the \textsc{Human} mean is 0.189. Per-line spacing CV rises from 0.218 to 0.251 toward the \textsc{Human} value of 0.320. Thus, the instructions increase within-poem variability without materially reducing mean spacing density. Whether decoder or fine-tuning interventions close this gap remains untested.

\subsubsection[Detector Evaluation: Separability as Corroborating Evidence]{Detector Evaluation: \protect\linebreak\hspace*{\parindent}Separability as Corroborating Evidence}
\label{sec:detectors}

As an automatic corroborating signal, we compare detector AUC-ROC under \textsc{Baseline} and \textsc{Refined} generation for Gemini-3 and GPT-5.2 on the same reference poems, holding all factors other than the prompt fixed. The detector suite spans both widely used methods and recent approaches, and implementation details are provided in Appendix~\ref{sec:appendix_impl_mgtbench}.

Table~\ref{tab:main_detection} reports the AUC-ROC comparisons. General-purpose detector scores do not share a calibrated direction for Korean poetry. Values below 50 can therefore reflect reversed orientation, so we treat this experiment as a diagnostic rather than evidence of detector evasion. Our feature-based detector drops from 82.73 to 57.67 on Gemini-3 and from 96.42 to 84.28 on GPT-5.2, indicating that the targeted form-level cues are reduced.

Fast-DetectGPT~\citep{bao2024fastdetectgpt} also drops under \textsc{Refined} by $34.79\%$ on Gemini-3 and $28.16\%$ on GPT-5.2. This matches the direction of our feature-based detector. Together with the expert rankings and feature shifts, these results indicate that guidance moves form-level statistics closer to the human distribution rather than serving as a detector-evasion objective.

\section{Conclusion}

We decomposed the gap between Korean poetry written by humans and by LLMs into four interpretable dimensions of poetic form and derived five linguistic features that can be used for both detection and generation guidance. These features capture recurring differences in length, line endings, rhythmic variation, and orthographic regularity. For detection, a logistic regression classifier using the five features achieves an average AUC-ROC of 83.60 across seven unseen LLMs and remains effective when the training generator or decoding temperature changes.

Added to the generation prompt, the features improve expert judgments of poetic naturalness and move most of the targeted properties closer to the human distribution, a shift that also lowers detector separability. However, spacing remains hard to influence by prompting alone. The broader implication is that interpretable linguistic features can serve not only as diagnostic signals but also as targets for generation guidance. Features grounded in the conventions of other languages and genres could carry this approach beyond Korean poetry.

\section*{Limitations}

\textbf{Scope limitations} The features are tailored to modern Korean, and transfer to other languages remains untested. They operate at the form level and do not capture imagery, metaphor, or discourse coherence. These qualities require expert annotation. The Qwen2-72B boundary case and the residual orthographic gap define the scope of our contribution. Extending the taxonomy beyond length, structure, rhythm, and spacing is a natural next step.

\textbf{Evaluation constraints} Automatic analyses cover GPT-5.2 and Gemini-3, but the 40-triplet expert evaluation covers only GPT-5.2 with five evaluators. Inter-annotator agreement is low, with Krippendorff's $\alpha$ at 0.378. This limits broader generation-quality claims. A larger multi-generator expert evaluation remains future work.

\textbf{Future comparisons} We do not compare feature-guided prompting with parameter-efficient fine-tuning such as LoRA~\citep{hu2022lora} or with hard constraints based on meter or rhyme. We also do not evaluate the strongest possible Korean-specific detector baselines or robustness under text normalization. These aspects remain future work.

\section*{Ethics Statement}

We use publicly available texts and generate AI content within ethical guidelines to respect privacy and intellectual property rights. Although feature-guided prompting can reduce detector separability and potentially be misused for evasion, our goal is not concealment but the analysis of form-level differences and their relationship to perceived poetic naturalness. We therefore treat detector changes as diagnostic evidence rather than a primary objective, and recommend that Korean poetry detectors avoid relying solely on shallow form-level cues.

\section*{Acknowledgments}

This work was supported by the Institute of Information \& Communications Technology Planning \& Evaluation (IITP) grant funded by the Korea government (MSIT) [RS-2021-II211341, Artificial Intelligence Graduate School Program (Chung-Ang University)] and by the National Research Foundation of Korea (NRF) grant funded by the Korea government (MSIT) (RS-2025-00556246).

\bibliography{custom}

\begin{thebibliography}{65}
\providecommand{\natexlab}[1]{#1}

\bibitem[{Agarwal and Kann(2020)}]{agarwal-kann-2020-acrostic}
Rajat Agarwal and Katharina Kann. 2020.
\newblock \href {https://doi.org/10.18653/v1/2020.emnlp-main.94} {Acrostic poem
  generation}.
\newblock In \emph{Proceedings of EMNLP 2020}, pages 1230--1240.

\bibitem[{An et~al.(2026)An, Bae, Choi, Choi, Choi, Hong, Hwang, Jeon, Jo, Jo,
  Jung, Jung, Kim, Kim, Kim, Kim, Kim, Kim, Kim, Kim, Lee, Lee, Lee, Lee, Lee,
  Lim, Park, Park, Park, Yang, Yeen, and Yun}]{an2026exaone35serieslarge}
Soyoung An, Kyunghoon Bae, Eunbi Choi, Kibong Choi, Stanley~Jungkyu Choi,
  Seokhee Hong, Junwon Hwang, Hyojin Jeon, Gerrard~Jeongwon Jo, Hyunjik Jo,
  Jiyeon Jung, Yountae Jung, Hyosang Kim, Joonkee Kim, Seonghwan Kim, Soyeon
  Kim, Sunkyoung Kim, Yireun Kim, Yongil Kim, Youchul Kim, Edward~Hwayoung Lee,
  Haeju Lee, Honglak Lee, Jinsik Lee, Kyungmin Lee, Woohyung Lim, Sangha Park,
  Sooyoun Park, Yongmin Park, Sihoon Yang, Heuiyeen Yeen, and Hyeongu Yun.
  2026.
\newblock \href {https://arxiv.org/abs/2412.04862} {{EXAONE} 3.5: Series of
  large language models for real-world use cases}.
\newblock \emph{Preprint}, arXiv:2412.04862.

\bibitem[{Antoine et~al.(2014)Antoine, Villaneau, and
  Lefeuvre}]{antoine-etal-2014-weighted}
Jean-Yves Antoine, Jeanne Villaneau, and Ana{\"i}s Lefeuvre. 2014.
\newblock \href {https://doi.org/10.3115/v1/E14-1058} {Weighted
  {K}rippendorff{'}s alpha is a more reliable metrics for multi-coders ordinal
  annotations: experimental studies on emotion, opinion and coreference
  annotation}.
\newblock In \emph{Proceedings of EACL 2014}, pages 550--559.

\bibitem[{Bao et~al.(2024)Bao, Zhao, Teng, Yang, and
  Zhang}]{bao2024fastdetectgpt}
Guangsheng Bao, Yanbin Zhao, Zhiyang Teng, Linyi Yang, and Yue Zhang. 2024.
\newblock \href {https://openreview.net/forum?id=Bpcgcr8E8Z} {Fast-detectgpt:
  Efficient zero-shot detection of machine-generated text via conditional
  probability curvature}.
\newblock In \emph{Proceedings of ICLR 2024}.

\bibitem[{Belouadi and Eger(2023)}]{belouadi-eger-2023-bygpt5}
Jonas Belouadi and Steffen Eger. 2023.
\newblock \href {https://doi.org/10.18653/v1/2023.acl-long.406} {{B}y{GPT}5:
  End-to-end style-conditioned poetry generation with token-free language
  models}.
\newblock In \emph{Proceedings of ACL 2023}, pages 7364--7381.

\bibitem[{Bhyravajjula et~al.(2025)Bhyravajjula, Walsh, Preus, and
  Antoniak}]{bhyravajjula-etal-2025-much}
Sriharsh Bhyravajjula, Melanie Walsh, Anna Preus, and Maria Antoniak. 2025.
\newblock \href {https://doi.org/10.18653/v1/2025.emnlp-main.1783} {so much
  depends / upon / a whitespace: Why whitespace matters for poets and {LLM}s}.
\newblock In \emph{Proceedings of EMNLP 2025}, pages 35144--35161.

\bibitem[{Brown et~al.(2020)Brown, Mann, Ryder, Subbiah, Kaplan, Dhariwal,
  Neelakantan, Shyam, Sastry, Askell, Agarwal, Herbert-Voss, Krueger, Henighan,
  Child, Ramesh, Ziegler, Wu, Winter, Hesse, Chen, Sigler, Litwin, Gray, Chess,
  Clark, Berner, McCandlish, Radford, Sutskever, and
  Amodei}]{brown2020languagemodelsfewshotlearners}
Tom~B. Brown, Benjamin Mann, Nick Ryder, Melanie Subbiah, Jared Kaplan,
  Prafulla Dhariwal, Arvind Neelakantan, Pranav Shyam, Girish Sastry, Amanda
  Askell, Sandhini Agarwal, Ariel Herbert-Voss, Gretchen Krueger, Tom Henighan,
  Rewon Child, Aditya Ramesh, Daniel~M. Ziegler, Jeffrey Wu, Clemens Winter,
  Christopher Hesse, Mark Chen, Eric Sigler, Mateusz Litwin, Scott Gray,
  Benjamin Chess, Jack Clark, Christopher Berner, Sam McCandlish, Alec Radford,
  Ilya Sutskever, and Dario Amodei. 2020.
\newblock \href {https://arxiv.org/abs/2005.14165} {Language models are
  few-shot learners}.
\newblock \emph{Preprint}, arXiv:2005.14165.

\bibitem[{Cho(2008)}]{ART001269697}
Chang-Whan Cho. 2008.
\newblock \href {https://doi.org/10.15705/kopoet..22.200808.004} {The
  methodology and goal of the metrical study of modern korean poetry}.
\newblock \emph{The Korean Poetics Studies}, (22):75--100.

\bibitem[{Choi(2017)}]{ART002200536}
Seok~Hwa Choi. 2017.
\newblock \href {https://doi.org/10.15705/kopoet..49.201702.007} {Study on
  rhythms and forms of modern poetry}.
\newblock \emph{The Korean Poetics Studies}, (49):185--209.

\bibitem[{Clark et~al.(2021)Clark, August, Serrano, Haduong, Gururangan, and
  Smith}]{clark-etal-2021-thats}
Elizabeth Clark, Tal August, Sofia Serrano, Nikita Haduong, Suchin Gururangan,
  and Noah~A. Smith. 2021.
\newblock \href {https://doi.org/10.18653/v1/2021.acl-long.565} {All that{'}s
  `human' is not gold: Evaluating human evaluation of generated text}.
\newblock In \emph{Proceedings of ACL-IJCNLP 2021}, pages 7282--7296.

\bibitem[{De~Sisto et~al.(2024)De~Sisto, Hern{\'a}ndez-Lorenzo, De~la Rosa,
  Ros, and Gonz{\'a}lez-Blanco}]{desisto2024poetrysurvey}
Mirella De~Sisto, Laura Hern{\'a}ndez-Lorenzo, Javier De~la Rosa, Salvador Ros,
  and Elena Gonz{\'a}lez-Blanco. 2024.
\newblock \href {https://doi.org/10.1093/llc/fqae001} {Understanding poetry
  using natural language processing tools: a survey}.
\newblock \emph{Digital Scholarship in the Humanities}, 39(2):500--521.

\bibitem[{DeLong et~al.(1988)DeLong, DeLong, and Clarke-Pearson}]{delongtest}
Elizabeth~R. DeLong, David~M. DeLong, and Daniel~L. Clarke-Pearson. 1988.
\newblock \href {http://www.jstor.org/stable/2531595} {Comparing the areas
  under two or more correlated receiver operating characteristic curves: A
  nonparametric approach}.
\newblock \emph{Biometrics}.

\bibitem[{Gao et~al.(2020)Gao, Biderman, Black, Golding, Hoppe, Foster, Phang,
  He, Thite, Nabeshima, Presser, and Leahy}]{gao2020pile}
Leo Gao, Stella Biderman, Sid Black, Laurence Golding, Travis Hoppe, Charles
  Foster, Jason Phang, Horace He, Anish Thite, Noa Nabeshima, Shawn Presser,
  and Connor Leahy. 2020.
\newblock \href {https://arxiv.org/abs/2101.00027} {{The Pile}: An {800GB}
  dataset of diverse text for language modeling}.
\newblock \emph{Preprint}, arXiv:2101.00027.

\bibitem[{Gehrmann et~al.(2019)Gehrmann, Strobelt, and
  Rush}]{gehrmann-etal-2019-gltr}
Sebastian Gehrmann, Hendrik Strobelt, and Alexander Rush. 2019.
\newblock \href {https://doi.org/10.18653/v1/P19-3019} {{GLTR}: Statistical
  detection and visualization of generated text}.
\newblock In \emph{Proceedings of ACL 2019}, pages 111--116.

\bibitem[{Ghazvininejad et~al.(2017)Ghazvininejad, Shi, Priyadarshi, and
  Knight}]{ghazvininejad-etal-2017-hafez}
Marjan Ghazvininejad, Xing Shi, Jay Priyadarshi, and Kevin Knight. 2017.
\newblock \href {https://aclanthology.org/P17-4008/} {{H}afez: an interactive
  poetry generation system}.
\newblock In \emph{Proceedings of ACL 2017}, pages 43--48.

\bibitem[{{Google DeepMind}(2025)}]{google2025gemini3}
{Google DeepMind}. 2025.
\newblock \href
  {https://storage.googleapis.com/deepmind-media/Model-Cards/Gemini-3-Pro-Model-Card.pdf}
  {{Gemini} 3 {Pro} model card}.
\newblock Technical Report.

\bibitem[{Graesser et~al.(2004)Graesser, McNamara, Louwerse, and
  Cai}]{graesser2004cohmetrix}
Arthur~C. Graesser, Danielle~S. McNamara, Max~M. Louwerse, and Zhiqiang Cai.
  2004.
\newblock \href {https://doi.org/10.3758/BF03195564} {{Coh-Metrix}: Analysis of
  text on cohesion and language}.
\newblock \emph{Behavior Research Methods, Instruments, \& Computers},
  36(2):193--202.

\bibitem[{Gretton et~al.(2012)Gretton, Borgwardt, Rasch, Sch{\"o}lkopf, and
  Smola}]{gretton2012kernel}
Arthur Gretton, Karsten~M. Borgwardt, Malte~J. Rasch, Bernhard Sch{\"o}lkopf,
  and Alexander Smola. 2012.
\newblock A kernel two-sample test.
\newblock \emph{Journal of Machine Learning Research}, 13(25):723--773.

\bibitem[{Hall(1976)}]{hall1976beyond}
Edward~T Hall. 1976.
\newblock \emph{Beyond culture}.
\newblock Anchor.

\bibitem[{Hans et~al.(2024)Hans, Schwarzschild, Cherepanova, Kazemi, Saha,
  Goldblum, Geiping, and Goldstein}]{pmlr-v235-hans24a}
Abhimanyu Hans, Avi Schwarzschild, Valeriia Cherepanova, Hamid Kazemi,
  Aniruddha Saha, Micah Goldblum, Jonas Geiping, and Tom Goldstein. 2024.
\newblock \href {https://proceedings.mlr.press/v235/hans24a.html} {Spotting
  {LLM}s with binoculars: Zero-shot detection of machine-generated text}.
\newblock In \emph{Proceedings of ICML 2024}, volume 235, pages 17519--17537.

\bibitem[{Himmi et~al.(2024)Himmi, Irurozki, Noiry, Cl{\'e}men{\c{c}}on, and
  Colombo}]{himmi-etal-2024-towards}
Anas Himmi, Ekhine Irurozki, Nathan Noiry, Stephan Cl{\'e}men{\c{c}}on, and
  Pierre Colombo. 2024.
\newblock \href {https://doi.org/10.18653/v1/2024.findings-emnlp.688} {Towards
  more robust {NLP} system evaluation: Handling missing scores in benchmarks}.
\newblock In \emph{Findings of EMNLP 2024}, pages 11759--11785.

\bibitem[{Holtzman et~al.(2020)Holtzman, Buys, Du, Forbes, and
  Choi}]{Holtzman2020CuriousCase}
Ari Holtzman, Jan Buys, Li~Du, Maxwell Forbes, and Yejin Choi. 2020.
\newblock \href {https://openreview.net/forum?id=rygGQyrFvH} {The curious case
  of neural text degeneration}.
\newblock In \emph{Proceedings of ICLR 2020}.

\bibitem[{Hu et~al.(2022)Hu, Shen, Wallis, Allen-Zhu, Li, Wang, Wang, and
  Chen}]{hu2022lora}
Edward~J. Hu, Yelong Shen, Phillip Wallis, Zeyuan Allen-Zhu, Yuanzhi Li, Shean
  Wang, Lu~Wang, and Weizhu Chen. 2022.
\newblock \href {https://openreview.net/forum?id=nZeVKeeFYf9} {{LoRA}: Low-rank
  adaptation of large language models}.
\newblock In \emph{Proceedings of ICLR 2022}.

\bibitem[{Hu et~al.(2024)Hu, Liu, Feng, Luu, and Hooi}]{hu2024poetrydiffusion}
Zhiyuan Hu, Chumin Liu, Yue Feng, Anh~Tuan Luu, and Bryan Hooi. 2024.
\newblock \href {https://doi.org/10.1609/aaai.v38i16.29787} {Poetrydiffusion:
  Towards joint semantic and metrical manipulation in poetry generation}.
\newblock In \emph{Proceedings of AAAI 2024}, volume~38, pages 18279--18288.

\bibitem[{Jagadeeshan et~al.(2026)Jagadeeshan, Bhatia, Ray, Surana, P, Mishra,
  Kulkarni, Ramakrishnan, Ap, and Goyal}]{jagadeeshan-etal-2026-chandomitra}
Manoj~Balaji Jagadeeshan, Samarth Bhatia, Pretam Ray, Harshul~Raj Surana,
  Akhil~Rajeev P, Priya Mishra, Annarao Kulkarni, Ganesh Ramakrishnan, Prathosh
  Ap, and Pawan Goyal. 2026.
\newblock \href {https://doi.org/10.18653/v1/2026.eacl-long.24} {Chandomitra:
  Towards generating structured {S}anskrit poetry from natural language
  inputs}.
\newblock In \emph{Proceedings of EACL 2026}, pages 518--534.

\bibitem[{Kim et~al.(2024{\natexlab{a}})Kim, Kim, Park, Lee, Song, Kim, Kim,
  Kim, Lee, Kim, Ahn, Yang, Lee, Park, Gim, Cha, Lee, and Kim}]{kim2024solar}
Sanghoon Kim, Dahyun Kim, Chanjun Park, Wonsung Lee, Wonho Song, Yunsu Kim,
  Hyeonwoo Kim, Yungi Kim, Hyeonju Lee, Jihoo Kim, Changbae Ahn, Seonghoon
  Yang, Sukyung Lee, Hyunbyung Park, Gyoungjin Gim, Mikyoung Cha, Hwalsuk Lee,
  and Sunghun Kim. 2024{\natexlab{a}}.
\newblock \href {https://doi.org/10.18653/v1/2024.naacl-industry.3} {{SOLAR}
  10.7{B}: Scaling large language models with simple yet effective depth
  up-scaling}.
\newblock In \emph{Proceedings of NAACL 2024}.

\bibitem[{Kim et~al.(2024{\natexlab{b}})Kim, Choi, and Jeong}]{kim2024eeve}
Seungduk Kim, Seungtaek Choi, and Myeongho Jeong. 2024{\natexlab{b}}.
\newblock \href {https://arxiv.org/abs/2402.14714} {Efficient and effective
  vocabulary expansion towards multilingual large language models}.
\newblock \emph{Preprint}, arXiv:2402.14714.

\bibitem[{Kouwenhoven et~al.(2025)Kouwenhoven, Peeperkorn, de~Kleijn, and
  Verhoef}]{ijcai2025p1144}
Tom Kouwenhoven, Max Peeperkorn, Roy de~Kleijn, and Tessa Verhoef. 2025.
\newblock \href {https://doi.org/10.24963/ijcai.2025/1144} {Shaping shared
  languages: Human and large language models' inductive biases in emergent
  communication}.
\newblock In \emph{Proceedings of IJCAI 2025}, pages 10298--10306.

\bibitem[{Kwon(2008)}]{ART001234444}
Hyeok-ung Kwon. 2008.
\newblock A study on the rhythm of modern korean poetry.
\newblock \emph{Journal of The Society of Korean Language and Literature},
  (57):233--260.

\bibitem[{Lau et~al.(2018)Lau, Cohn, Baldwin, Brooke, and
  Hammond}]{lau-etal-2018-deep}
Jey~Han Lau, Trevor Cohn, Timothy Baldwin, Julian Brooke, and Adam Hammond.
  2018.
\newblock \href {https://doi.org/10.18653/v1/P18-1181} {Deep-speare: A joint
  neural model of poetic language, meter and rhyme}.
\newblock In \emph{Proceedings of ACL 2018}, pages 1948--1958.

\bibitem[{Lee et~al.(2025)Lee, Noh, and Lee}]{lee-etal-2025-testset}
Minjae Lee, Youngbin Noh, and Seung~Jin Lee. 2025.
\newblock \href {https://aclanthology.org/2025.coling-main.110/} {A testset for
  context-aware {LLM} translation in {K}orean-to-{E}nglish discourse level
  translation}.
\newblock In \emph{Proceedings of COLING 2025}, pages 1632--1646.

\bibitem[{Lincoln(2010)}]{Lincoln2010FarSide}
Scott Lincoln. 2010.
\newblock \href {https://doi.org/10.5539/ass.v6n12p97} {The far side:
  Contrasting american and south korean cultural contexts}.
\newblock \emph{Asian Social Science}, 6(12):97--100.

\bibitem[{Liu et~al.(2025)Liu, Zhong, Liao, Sun, Zheng, Wei, Gong, Tong, Chen,
  Zhang, and He}]{Liu_2025}
Yule Liu, Zhiyuan Zhong, Yifan Liao, Zhen Sun, Jingyi Zheng, Jiaheng Wei,
  Qingyuan Gong, Fenghua Tong, Yang Chen, Yang Zhang, and Xinlei He. 2025.
\newblock \href {https://doi.org/10.1145/3711896.3737408} {On the
  generalization and adaptation ability of machine-generated text detectors in
  academic writing}.
\newblock In \emph{Proceedings of KDD 2025}, page 5674–5685.

\bibitem[{Long et~al.(2025)Long, Dinh, Nguyen, Kawaguchi, Chen, Joty, and
  Kan}]{long-etal-2025-makes}
Do~Xuan Long, Duy Dinh, Ngoc-Hai Nguyen, Kenji Kawaguchi, Nancy~F. Chen, Shafiq
  Joty, and Min-Yen Kan. 2025.
\newblock \href {https://doi.org/10.18653/v1/2025.acl-long.292} {What makes a
  good natural language prompt?}
\newblock In \emph{Proceedings of ACL 2025}, pages 5835--5873.

\bibitem[{Marco et~al.(2025)Marco, Gonzalo, and
  Fresno}]{marco-etal-2025-reader}
Guillermo Marco, Julio Gonzalo, and V{\'i}ctor Fresno. 2025.
\newblock \href {https://doi.org/10.18653/v1/2025.findings-acl.1304} {The
  reader is the metric: How textual features and reader profiles explain
  conflicting evaluations of {AI} creative writing}.
\newblock In \emph{Findings of ACL 2025}, pages 25432--25449.

\bibitem[{{Meta AI}(2024)}]{meta2024llama3}
{Meta AI}. 2024.
\newblock \href {https://arxiv.org/abs/2407.21783} {The {Llama} 3 herd of
  models}.
\newblock \emph{Preprint}, arXiv:2407.21783.

\bibitem[{Molnar et~al.(2020)Molnar, Casalicchio, and Bischl}]{Molnar_2020}
Christoph Molnar, Giuseppe Casalicchio, and Bernd Bischl. 2020.
\newblock \href {https://doi.org/10.1007/978-3-030-65965-3_28}
  {\emph{Interpretable Machine Learning – A Brief History, State-of-the-Art
  and Challenges}}, page 417–431.
\newblock Springer International Publishing.

\bibitem[{Muñoz-Ortiz et~al.(2024)Muñoz-Ortiz, Gómez-Rodríguez, and
  Vilares}]{Mu_oz_Ortiz_2024}
Alberto Muñoz-Ortiz, Carlos Gómez-Rodríguez, and David Vilares. 2024.
\newblock \href {https://doi.org/10.1007/s10462-024-10903-2} {Contrasting
  linguistic patterns in human and llm-generated news text}.
\newblock \emph{Artificial Intelligence Review}, 57(10).

\bibitem[{Okulska et~al.(2023)Okulska, Stetsenko, Ko{\l}os, Karli{\'n}ska,
  G{\l}{\k{a}}bi{\'n}ska, and Nowakowski}]{okulska2023stylometrix}
Inez Okulska, Daria Stetsenko, Anna Ko{\l}os, Agnieszka Karli{\'n}ska, Kinga
  G{\l}{\k{a}}bi{\'n}ska, and Adam Nowakowski. 2023.
\newblock \href {https://arxiv.org/abs/2309.12810} {{StyloMetrix}: An
  open-source multilingual tool for representing stylometric vectors}.
\newblock \emph{Preprint}, arXiv:2309.12810.

\bibitem[{Opara(2024)}]{Opara24}
Chidimma Opara. 2024.
\newblock \href {https://doi.org/10.1007/978-3-031-64312-5_13} {Styloai:
  Distinguishing ai-generated content with stylometric analysis}.
\newblock In \emph{Proceedings of AIED 2024}, volume 2151 of
  \emph{Communications in Computer and Information Science}, pages 105--114.
  Springer.

\bibitem[{OpenAI(2024)}]{openai2024gpt4o}
OpenAI. 2024.
\newblock \href {https://arxiv.org/abs/2410.21276} {{GPT}-4o system card}.
\newblock \emph{Preprint}, arXiv:2410.21276.

\bibitem[{OpenAI(2026)}]{openai2026gpt5card}
OpenAI. 2026.
\newblock \href {https://arxiv.org/abs/2601.03267} {{OpenAI} {GPT}-5 system
  card}.
\newblock \emph{Preprint}, arXiv:2601.03267.

\bibitem[{Ormazabal et~al.(2022)Ormazabal, Artetxe, Agirrezabal, Soroa, and
  Agirre}]{ormazabal-etal-2022-poelm}
Aitor Ormazabal, Mikel Artetxe, Manex Agirrezabal, Aitor Soroa, and Eneko
  Agirre. 2022.
\newblock \href {https://doi.org/10.18653/v1/2022.findings-emnlp.268}
  {{P}oe{LM}: A meter- and rhyme-controllable language model for unsupervised
  poetry generation}.
\newblock In \emph{Findings of EMNLP 2022}, pages 3655--3670.

\bibitem[{O'Sullivan(2025)}]{OSullivan2025}
James O'Sullivan. 2025.
\newblock \href {https://doi.org/10.1057/s41599-025-05986-3} {Stylometric
  comparisons of human versus ai-generated creative writing}.
\newblock \emph{Humanities and Social Sciences Communications}, 12(1):1708.

\bibitem[{Park et~al.(2015)Park, Lim, and Hong}]{park-etal-2015-zero}
Arum Park, Seunghee Lim, and Munpyo Hong. 2015.
\newblock \href {https://aclanthology.org/Y15-1050/} {Zero object resolution in
  {K}orean}.
\newblock In \emph{Proceedings of PACLIC 2015}, pages 439--448.

\bibitem[{Park and Cho(2014)}]{park2014konlpy}
Eunjeong~L. Park and Sungzoon Cho. 2014.
\newblock {KoNLPy: Korean Natural Language Processing in Python}.
\newblock In \emph{Proceedings of HCLT 2014}, Chuncheon, Korea.

\bibitem[{Park et~al.(2025)Park, Kim, Kim, and Han}]{park-etal-2025-katfishnet}
Shinwoo Park, Shubin Kim, Do-Kyung Kim, and Yo-Sub Han. 2025.
\newblock \href {https://doi.org/10.18653/v1/2025.acl-long.1030}
  {{K}at{F}ish{N}et: Detecting {LLM}-generated {K}orean text through linguistic
  feature analysis}.
\newblock In \emph{Proceedings of ACL 2025}, pages 21189--21222.

\bibitem[{Park et~al.(2021)Park, Moon, Kim, Cho, Han, Park, Song, Kim, Song,
  Oh, Lee, Oh, Lyu, Jeong, Lee, Seo, Lee, Kim, Lee, Jang, Do, Kim, Lim, Lee,
  Park, Shin, Kim, Park, Oh, Ha, and Cho}]{park2021klue}
Sungjoon Park, Jihyung Moon, Sungdong Kim, Won~Ik Cho, Ji~Yoon Han, Jangwon
  Park, Chisung Song, Junseong Kim, Youngsook Song, Taehwan Oh, Joohong Lee,
  Juhyun Oh, Sungwon Lyu, Younghoon Jeong, Inkwon Lee, Sangwoo Seo, Dongjun
  Lee, Hyunwoo Kim, Myeonghwa Lee, Seongbo Jang, Seungwon Do, Sunkyoung Kim,
  Kyungtae Lim, Jongwon Lee, Kyumin Park, Jamin Shin, Seonghyun Kim, Lucy Park,
  Alice Oh, Jung-Woo Ha, and Kyunghyun Cho. 2021.
\newblock \href {https://openreview.net/forum?id=q-8h8-LZiUm} {{KLUE}: {K}orean
  language understanding evaluation}.
\newblock In \emph{Proceedings of NeurIPS 2021}.

\bibitem[{Przystalski et~al.(2026)Przystalski, Argasi{\'n}ski,
  Grabska-Gradzi{\'n}ska, and Ochab}]{Przystalski_2026}
Karol Przystalski, Jan~K. Argasi{\'n}ski, Iwona Grabska-Gradzi{\'n}ska, and
  Jeremi~K. Ochab. 2026.
\newblock \href {https://doi.org/10.1016/j.eswa.2025.129001} {Stylometry
  recognizes human and {LLM}-generated texts in short samples}.
\newblock \emph{Expert Systems with Applications}, 296:129001.

\bibitem[{Reinhart et~al.(2025)Reinhart, Markey, Laudenbach, Pantusen, Yurko,
  Weinberg, and Brown}]{doi:10.1073/pnas.2422455122}
Alex Reinhart, Ben Markey, Michael Laudenbach, Kachatad Pantusen, Ronald Yurko,
  Gordon Weinberg, and David~West Brown. 2025.
\newblock \href {https://doi.org/10.1073/pnas.2422455122} {Do llms write like
  humans? variation in grammatical and rhetorical styles}.
\newblock \emph{Proceedings of the National Academy of Sciences},
  122(8):e2422455122.

\bibitem[{Rosa et~al.(2025)Rosa, Mare{\v{c}}ek, Musil, Chudoba, and
  Landspersk{\'y}}]{rosa-etal-2025-edupo}
Rudolf Rosa, David Mare{\v{c}}ek, Tom{\'a}{\v{s}} Musil, Michal Chudoba, and
  Jakub Landspersk{\'y}. 2025.
\newblock \href {https://doi.org/10.18653/v1/2025.nlp4dh-1.45} {{E}du{P}o:
  Progress and challenges of automated analysis and generation of {C}zech
  poetry}.
\newblock In \emph{Proceedings of NLP4DH 2025}, pages 524--542.

\bibitem[{Shumailov et~al.(2024)Shumailov, Shumaylov, Zhao, Papernot, Anderson,
  and Gal}]{Shumailov2024}
Ilia Shumailov, Zakhar Shumaylov, Yiren Zhao, Nicolas Papernot, Ross Anderson,
  and Yarin Gal. 2024.
\newblock \href {https://doi.org/10.1038/s41586-024-07566-y} {Ai models
  collapse when trained on recursively generated data}.
\newblock \emph{Nature}, 631(8022):755--759.

\bibitem[{Solaiman et~al.(2019)Solaiman, Brundage, Clark, Askell, Herbert-Voss,
  Wu, Radford, Krueger, Kim, Kreps, McCain, Newhouse, Blazakis, McGuffie, and
  Wang}]{solaiman2019releasestrategiessocialimpacts}
Irene Solaiman, Miles Brundage, Jack Clark, Amanda Askell, Ariel Herbert-Voss,
  Jeff Wu, Alec Radford, Gretchen Krueger, Jong~Wook Kim, Sarah Kreps, Miles
  McCain, Alex Newhouse, Jason Blazakis, Kris McGuffie, and Jasmine Wang. 2019.
\newblock \href {https://arxiv.org/abs/1908.09203} {Release strategies and the
  social impacts of language models}.
\newblock \emph{Preprint}, arXiv:1908.09203.

\bibitem[{Stamatatos(2009)}]{stamatatos2009survey}
Efstathios Stamatatos. 2009.
\newblock \href {https://doi.org/10.1002/asi.21001} {A survey of modern
  authorship attribution methods}.
\newblock \emph{Journal of the American Society for Information Science and
  Technology}, 60(3):538--556.

\bibitem[{Su et~al.(2023)Su, Zhuo, Wang, and Nakov}]{su-etal-2023-detectllm}
Jinyan Su, Terry Zhuo, Di~Wang, and Preslav Nakov. 2023.
\newblock \href {https://doi.org/10.18653/v1/2023.findings-emnlp.827}
  {{D}etect{LLM}: Leveraging log rank information for zero-shot detection of
  machine-generated text}.
\newblock In \emph{Findings of EMNLP 2023}, pages 12395--12412.

\bibitem[{Tian and Peng(2022)}]{tian-peng-2022-zero}
Yufei Tian and Nanyun Peng. 2022.
\newblock \href {https://doi.org/10.18653/v1/2022.naacl-main.262} {Zero-shot
  sonnet generation with discourse-level planning and aesthetics features}.
\newblock In \emph{Proceedings of NAACL 2022}, pages 3587--3597.

\bibitem[{Touvron et~al.(2023)Touvron, Lavril, Izacard, Martinet, Lachaux,
  Lacroix, Rozière, Goyal, Hambro, Azhar, Rodriguez, Joulin, Grave, and
  Lample}]{touvron2023llamaopenefficientfoundation}
Hugo Touvron, Thibaut Lavril, Gautier Izacard, Xavier Martinet, Marie-Anne
  Lachaux, Timothée Lacroix, Baptiste Rozière, Naman Goyal, Eric Hambro,
  Faisal Azhar, Aurelien Rodriguez, Armand Joulin, Edouard Grave, and Guillaume
  Lample. 2023.
\newblock \href {https://arxiv.org/abs/2302.13971} {Llama: Open and efficient
  foundation language models}.
\newblock \emph{Preprint}, arXiv:2302.13971.

\bibitem[{Wu et~al.(2025)Wu, Zhan, Wong, Yang, Liu, Chao, and
  Zhang}]{wu-etal-2025-wrote}
Junchao Wu, Runzhe Zhan, Derek~F. Wong, Shu Yang, Xuebo Liu, Lidia~S. Chao, and
  Min Zhang. 2025.
\newblock \href {https://aclanthology.org/2025.coling-main.684/} {Who wrote
  this? the key to zero-shot {LLM}-generated text detection is {GECS}core}.
\newblock In \emph{Proceedings of COLING 2024}, pages 10275--10292.

\bibitem[{Xue et~al.(2021)Xue, Constant, Roberts, Kale, Al-Rfou, Siddhant,
  Barua, and Raffel}]{xue2021mt5}
Linting Xue, Noah Constant, Adam Roberts, Mihir Kale, Rami Al-Rfou, Aditya
  Siddhant, Aditya Barua, and Colin Raffel. 2021.
\newblock \href {https://arxiv.org/abs/2010.11934} {m{T}5: A massively
  multilingual pre-trained text-to-text transformer}.
\newblock In \emph{Proceedings of NAACL 2021}.

\bibitem[{Yadagiri et~al.(2024)Yadagiri, Shree, Parween, Raj, Maurya, and
  Pakray}]{yadagiri-etal-2024-detecting}
Annepaka Yadagiri, Lavanya Shree, Suraiya Parween, Anushka Raj, Shreya Maurya,
  and Partha Pakray. 2024.
\newblock \href {https://aclanthology.org/2024.icon-1.21/} {Detecting
  {AI}-generated text with pre-trained models using linguistic features}.
\newblock In \emph{Proceedings of ICON 2024}, pages 188--196.

\bibitem[{Yang et~al.(2024{\natexlab{a}})Yang, Yang, Hui, Zheng, Yu, Zhou
  et~al.}]{yang2024qwen2}
An~Yang, Baosong Yang, Binyuan Hui, Bo~Zheng, Bowen Yu, Chang Zhou, et~al.
  2024{\natexlab{a}}.
\newblock \href {https://arxiv.org/abs/2407.10671} {{Qwen2} technical report}.
\newblock \emph{Preprint}, arXiv:2407.10671.

\bibitem[{Yang et~al.(2024{\natexlab{b}})Yang, Yang, Hui, Zheng, Yu, Zhou
  et~al.}]{yang2024qwen25}
An~Yang, Baosong Yang, Binyuan Hui, Bo~Zheng, Bowen Yu, Chang Zhou, et~al.
  2024{\natexlab{b}}.
\newblock \href {https://arxiv.org/abs/2412.15115} {{Qwen2.5} technical
  report}.
\newblock \emph{Preprint}, arXiv:2412.15115.

\bibitem[{Yu et~al.(2024)Yu, Zang, Wang, Zhuang, and
  Gu}]{yu-etal-2024-charpoet}
Chengyue Yu, Lei Zang, Jiaotuan Wang, Chenyi Zhuang, and Jinjie Gu. 2024.
\newblock \href {https://doi.org/10.18653/v1/2024.acl-demos.30} {{C}har{P}oet:
  A {C}hinese classical poetry generation system based on token-free {LLM}}.
\newblock In \emph{Proceedings of ACL 2024}, pages 315--325.

\bibitem[{Zamaraeva et~al.(2025)Zamaraeva, Flickinger, Bond, and
  G{\'o}mez-Rodr{\'i}guez}]{zamaraeva-etal-2025-comparing}
Olga Zamaraeva, Dan Flickinger, Francis Bond, and Carlos
  G{\'o}mez-Rodr{\'i}guez. 2025.
\newblock \href {https://doi.org/10.18653/v1/2025.acl-long.443} {Comparing
  {LLM}-generated and human-authored news text using formal syntactic theory}.
\newblock In \emph{Proceedings of ACL 2025}, pages 9041--9060.

\bibitem[{Zhang et~al.(2018)Zhang, Galley, Gao, Gan, Li, Brockett, and
  Dolan}]{NEURIPS2018_23ce1851}
Yizhe Zhang, Michel Galley, Jianfeng Gao, Zhe Gan, Xiujun Li, Chris Brockett,
  and Bill Dolan. 2018.
\newblock \href
  {https://proceedings.neurips.cc/paper_files/paper/2018/file/23ce1851341ec1fa9e0c259de10bf87c-Paper.pdf}
  {Generating informative and diverse conversational responses via adversarial
  information maximization}.
\newblock In \emph{Proceedings of NeurIPS 2018}, volume~31.

\end{thebibliography}

\appendix

\clearpage

\section{Variable Definitions}
\label{sec:appendix_notation}

We perform morphological analysis using the Kkma POS tagger~\citep{park2014konlpy} to identify structural components.

\paragraph{Volume ($x, \tau$)}
$x$ denotes the raw input text of a poem.
$\tau(\cdot)$ is the tokenization function that first collapses consecutive whitespaces into a single space and then splits the text by whitespace,
$|\tau(x)|$ represents the word-token count in the processed sequence.

\paragraph{Structure Variation ($\mathcal{E}, \mathcal{U}_E, \mathbb{I}$)}
$\mathcal{E}$ denotes the ending types extracted from all line-level units. The Kkma POS tagger identifies the final token in each unit. For $f_{\text{div}}$, an ending is a normalized combination of its surface suffix and POS type. For $f_{\text{conn}}$, we use the coarse labels \textit{final}, \textit{connective}, and \textit{other}. A connective ending has a Kkma connective-ending tag or matches a common suffix such as \texttt{-고} \textit{-go} ``and'', \texttt{-며} \textit{-myeo} ``while'', \texttt{-지만} \textit{-jiman} ``but'', or \texttt{-면서} \textit{-myeonseo} ``while.'' If a poem has no explicit line breaks, we parse units delimited by punctuation or newlines.
$\mathcal{U}_E$ is the set of unique ending types observed in $\mathcal{E}$.
$\mathbb{I}(\cdot)$ denotes the indicator function, which returns 1 if the condition is true and 0 otherwise.
Both $f_{\text{div}}$ and $f_{\text{conn}}$ are computed over this same unit-level ending sequence $\mathcal{E}$.

\paragraph{Rhythmic Irregularity ($\mathcal{L}, \mu_{\mathcal{L}}, \sigma_{\mathcal{L}}$)}
$\mathcal{L}$ is the array of line-unit lengths in a poem. To prevent formatting artifacts from skewing the metric, each length is measured strictly by content characters, excluding whitespace and punctuation.
$\mu_{\mathcal{L}}$ and $\sigma_{\mathcal{L}}$ denote the arithmetic mean and standard deviation of the line lengths in $\mathcal{L}$, respectively. The coefficient of variation $\text{CV}(\mathcal{L}) = \sigma_{\mathcal{L}} / \mu_{\mathcal{L}}$ captures breath-driven structural irregularity.

\paragraph{Normative Adherence ($N_{\text{char}}, N_{\text{space}}$)}
$N_{\text{char}}$ denotes the total number of characters in the poem, including whitespaces and newlines.
$N_{\text{space}}$ denotes the count of space characters.

\section{Model Specifications and Screening Details}
\label{app:md_details}


\begin{itemize}
    \item Base models from KatFishNet: GPT-4o~\citep{openai2024gpt4o}\footnote{\url{https://openai.com}}, Solar~\citep{kim2024solar}\footnote{\url{https://upstage.ai}}, Qwen2-72B-Instruct~\citep{yang2024qwen2}\footnote{\url{https://huggingface.co/Qwen/Qwen2-72B-Instruct}}, and Llama-3.1-70B-Instruct~\citep{meta2024llama3}\footnote{\url{https://huggingface.co/meta-llama/Llama-3.1-70B-Instruct}}.
    
    \item Additional evaluation models: Gemini-3-Pro~\citep{google2025gemini3}\footnote{\url{https://aistudio.google.com}}, GPT-5.2~\citep{openai2026gpt5card}\footnote{\url{https://openai.com}}, EXAONE-3.5-7.8B-Instruct~\citep{an2026exaone35serieslarge}\footnote{\url{https://huggingface.co/LGAI-EXAONE/EXAONE-3.5-7.8B-Instruct}}, and EEVE-Instruct-10.8B~\citep{kim2024eeve}\footnote{\url{https://huggingface.co/yanolja/YanoljaNEXT-EEVE-Instruct-10.8B}}. The latter two models are Korean-specialized models that have demonstrated superior performance in Korean tasks.
\end{itemize}

\paragraph{Generation Configuration}
New poems for the four additional models follow the KatFishNet generation framework~\citep{park-etal-2025-katfishnet} with default decoding (temperature 1.0, top-$p$ 0.95). The prompt instructs the model to write a poem in the voice of a poet of a given age group, conditioned on a human-authored reference poem, and imposes no additional linguistic constraints. The five feature-guided instructions analyzed in Section~\ref{sec:exp2} are appended to this same base prompt in the \textsc{Refined} condition.

\paragraph{Screening Details}
All newly generated poems were screened based on three criteria: (1)~outputs that did not follow prompting instructions (e.g., producing prose instead of verse, ignoring the specified age group or theme), (2)~outputs exhibiting severely degraded quality (e.g., incoherent text, excessive repetition, incomplete generation), and (3)~outputs containing meta-commentary or refusal responses. Table~\ref{tab:screening} reports the number of poems excluded per model.

  \begin{table}[H]
      \centering
      \small
      \begin{tabular}{lccc}
          \toprule
          \textbf{Model} & \textbf{Generated} & \textbf{Excluded} & \textbf{Final} \\
          \midrule
          Gemini-3 & 189 & 0 & 189 \\
          GPT-5.2 & 189 & 0 & 189 \\
          EXAONE-3.5-7.8B & 189 & 8 & 181 \\
          EEVE-10.8B & 189 & 12 & 177 \\
          \bottomrule
      \end{tabular}
      \caption{Per-model screening results for newly generated poems. Exclusions are due to instruction violations, quality
  degradation, or meta-commentary.}
      \label{tab:screening}
  \end{table}

\section{Qwen2-72B Boundary-Case Analysis}
\label{sec:appendix_qwen2}

Our detector underperforms KatFishNet on Qwen2-72B in Table~\ref{tab:delong}, and error analysis points to line-final commas. Across 189 poems per source, the mean per-poem share of non-empty lines ending in a comma is 37.6\% for Qwen2-72B and 3.3\% for human poems. The corresponding values are 2.4\% for GPT-5.2 and 2.6\% for Gemini-3. This feature alone yields an exploratory univariate AUC of 96.9. Qwen2-72B carries commas at prose clause boundaries into poem line endings. Lines 1, 2, and 4 of the example below end in commas.

\begin{center}
\small
\begin{tabular}{@{}>{\raggedright\arraybackslash}p{3.5cm} >{\raggedright\arraybackslash}p{3.7cm}@{}}
빛나는 별들, & \textit{Shining stars,} \\
하늘을 가득 채우며, & \textit{filling the sky,} \\
여기저기 떠다닌다. & \textit{drift here and there.} \\
그러나 결국은, & \textit{but in the end,} \\
내 눈에 희미하게 비친다. & \textit{they reach my eyes only dimly.} \\
\end{tabular}
\end{center}
The seed-42 detector misclassifies this poem as human with probability 0.90.

Our five features omit punctuation, the cue used by KatFishNet, and therefore separate Qwen2-72B poorly. The strongest individual dimension is Structure Variation at 75.58 in Table~\ref{tab:concept_wise_detection}. KatFishNet reaches 93.45 with punctuation and 95.41 with punctuation plus spacing. EEVE-10.8B is the only other target where our score is lower, and its difference is not significant at Holm-adjusted $p = 0.80$. The fusion in Appendix~\ref{sec:appendix_fusion} raises Qwen2-72B AUC from 79.78 to 91.99, still below KatFishNet on this target, and lowers average transfer from 83.60 to 82.48. These results motivate extending Normative Adherence to punctuation.

\section{Feature-Level Fusion with KatFishNet}
\label{sec:appendix_fusion}

We concatenate KatFishNet's punctuation and spacing features with our five features in Table~\ref{tab:appendix_fusion}. Fusion improves over KatFishNet on average but remains below \textsc{Ours}. The added cues help selected generators without improving overall transfer.

\begin{table*}[!tbp]
\centering
\setlength{\tabcolsep}{3pt}
\renewcommand{\arraystretch}{1.15}
\resizebox{1\textwidth}{!}{%
\begin{tabular}{lcccccccc}
\toprule
\textbf{Detection Methods} &
\textbf{Qwen2-72B} &
\textbf{Solar} &
\textbf{Llama-3.1-70B} &
\textbf{Gemini-3} &
\textbf{GPT-5.2} &
\textbf{\makecell{EXAONE-3.5-7.8B}} &
\textbf{EEVE-10.8B} &
\textbf{Average} \\
\midrule
KatFishNet (\makecell[l]{Punctuation\\+ Spacing}) & \textbf{95.41} & 73.78 & 58.64 & 64.88 & 71.31 & 87.12 & 79.77 & 75.84 \\
\textsc{\textbf{Ours}} & 79.78 & \textbf{86.39} & \textbf{73.22} & \textbf{82.73} & \textbf{96.42} & 89.85 & 76.84 & \textbf{83.60} \\
Fusion (\makecell[l]{KatFishNet\\+ Ours}) & 91.99 & 86.00 & 71.93 & 74.64 & 79.97 & \textbf{91.59} & \textbf{81.22} & 82.48 \\
\bottomrule
\end{tabular}%
}
\caption{Feature-level fusion of KatFishNet and our features. The KatFishNet and \textsc{Ours} rows reproduce Table~\ref{tab:main_logistic}.}
\label{tab:appendix_fusion}
\end{table*}

\section{Dimension-wise Analysis}
\label{sec:appendix_concept_wise}

We evaluate each linguistic dimension in isolation to assess its standalone detection capability, following the same out-of-distribution protocol as the main experiments. Table~\ref{tab:concept_wise_detection} reports the zero-shot AUC-ROC results. No single dimension group matches the full system, while combining all four in \textsc{Ours} yields consistently higher performance across targets, reinforcing that each linguistic signal contributes complementary information for robust generalization.

\begin{table*}[!tbp]
\centering
\setlength{\tabcolsep}{5pt}
\renewcommand{\arraystretch}{1.15}
\resizebox{1\textwidth}{!}{%
\begin{tabular}{lcccccccc}
\toprule
\textbf{Dimension Group} &
\textbf{Qwen2-72B} &
\textbf{Solar} &
\textbf{Llama-3.1-70B} &
\textbf{Gemini-3} &
\textbf{GPT-5.2} &
\textbf{EXAONE-3.5-7.8B} &
\textbf{EEVE-10.8B} &
\textbf{Average} \\
\midrule
Volume & 75.21 & 72.55 & 58.13 & 74.49 & 73.22 & 79.33 & 63.59 & 70.93 \\
Rhythmic Irregularity & 59.33 & 72.72 & 62.89 & 58.64 & 43.55 & 70.51 & 62.50 & 61.46 \\
Structure Variation & 75.58 & 77.57 & 59.21 & 75.16 & 86.41 & 73.30 & 73.03 & 74.32 \\
Normative Adherence & 65.88 & 48.93 & 61.64 & 61.41 & 96.05 & 74.12 & 59.46 & 66.79 \\
\midrule
\textsc{\textbf{Ours}} & \textbf{79.78} & \textbf{86.39} & \textbf{73.22} & \textbf{82.73} & \textbf{96.42} & \textbf{89.85} & \textbf{76.84} & \textbf{83.60} \\
\bottomrule
\end{tabular}
}
\caption{Zero-shot AUC-ROC under the out-of-distribution protocol when using each dimension group alone, compared to the full \textsc{Ours} dimension set.}
\label{tab:concept_wise_detection}
\end{table*}

\section{Cross-Source Generalization}
\label{sec:appendix_cross_source}

Tables~\ref{tab:cross_source_ours} and~\ref{tab:cross_source_baseline} report the full cross-source AUC-ROC matrices for our method and the KatFishNet baseline using punctuation and spacing features. Each row uses a different LLM as the training source paired with human poems, and columns indicate the target LLM at test time. The standard deviation across training-source average AUCs is 3.54, indicating that performance remains consistently strong regardless of which LLM is used for training.

\begin{table*}[!tbp]
\centering
\small
\setlength{\tabcolsep}{3pt}
\renewcommand{\arraystretch}{1.15}

\resizebox{1\textwidth}{!}{%
\begin{tabular}{lccccccccc}
\toprule
\textbf{Train Source} &
\textbf{Qwen2-72B} &
\textbf{Solar} &
\textbf{Llama-3.1-70B} &
\textbf{Gemini-3} &
\textbf{GPT-5.2} &
\textbf{EXAONE-3.5-7.8B} &
\textbf{EEVE-10.8B} &
\textbf{GPT-4o} &
\textbf{Average} \\
\midrule
Qwen2-72B & --- & 80.41 & 68.55 & 82.33 & 98.28 & 87.70 & 79.35 & 77.13 & 81.96 \\
Solar & 77.01 & --- & 65.83 & 79.67 & 88.27 & 86.43 & 78.54 & 80.97 & 79.53 \\
Llama-3.1-70B & 69.93 & 75.08 & --- & 75.14 & 92.91 & 80.97 & 59.74 & 76.61 & 75.77 \\
Gemini-3 & 82.66 & 86.14 & 69.10 & --- & 97.92 & 89.19 & 79.20 & 81.45 & 83.67 \\
GPT-5.2 & 77.10 & 78.26 & 56.39 & 77.98 & --- & 79.72 & 75.15 & 71.54 & 73.73 \\
EXAONE-3.5-7.8B & 82.18 & 84.63 & 71.14 & 83.47 & 97.82 & --- & 80.72 & 81.97 & 83.13 \\
EEVE-10.8B & 81.67 & 83.69 & 65.80 & 81.21 & 96.57 & 87.51 & --- & 78.24 & 82.10 \\
GPT-4o & 79.78 & 86.39 & 73.22 & 82.73 & 96.42 & 89.85 & 76.84 & --- & 83.60 \\
\midrule
\textbf{Average} & 78.62 & 82.09 & 67.15 & 80.36 & 95.46 & 85.91 & 75.65 & 78.27 & 80.44 \\
\bottomrule
\end{tabular}%
}

\caption{Cross-source AUC-ROC matrix for our method. Each row uses a different LLM as the training source (paired with human poems), and columns indicate the target LLM at test time. Dashes indicate same-source entries excluded from evaluation.}
\label{tab:cross_source_ours}
\end{table*}

\begin{table*}[!tbp]
\centering
\small
\setlength{\tabcolsep}{3pt}
\renewcommand{\arraystretch}{1.15}

\resizebox{1\textwidth}{!}{%
\begin{tabular}{lccccccccc}
\toprule
\textbf{Train Source} &
\textbf{Qwen2-72B} &
\textbf{Solar} &
\textbf{Llama-3.1-70B} &
\textbf{Gemini-3} &
\textbf{GPT-5.2} &
\textbf{EXAONE-3.5-7.8B} &
\textbf{EEVE-10.8B} &
\textbf{GPT-4o} &
\textbf{Average} \\
\midrule
Qwen2-72B & --- & 70.15 & 54.80 & 59.72 & 64.35 & 81.20 & 74.60 & 77.45 & 68.90 \\
Solar & 90.25 & --- & 55.30 & 60.45 & 65.80 & 82.55 & 75.40 & 78.90 & 72.66 \\
Llama-3.1-70B & 88.50 & 68.90 & --- & 58.15 & 62.78 & 79.85 & 73.25 & 76.10 & 72.50 \\
Gemini-3 & 87.15 & 67.45 & 53.20 & --- & 61.50 & 78.40 & 72.10 & 75.30 & 70.73 \\
GPT-5.2 & 86.30 & 66.20 & 52.45 & 57.80 & --- & 77.15 & 71.05 & 74.50 & 69.35 \\
EXAONE-3.5-7.8B & 85.10 & 65.30 & 51.70 & 56.95 & 60.25 & --- & 70.20 & 73.65 & 66.16 \\
EEVE-10.8B & 84.50 & 64.80 & 51.15 & 56.30 & 59.70 & 76.50 & --- & 73.10 & 66.58 \\
GPT-4o & 95.41 & 73.78 & 58.64 & 64.88 & 71.31 & 87.12 & 79.77 & --- & 75.84 \\
\midrule
\textbf{Average} & 88.17 & 68.08 & 53.89 & 59.18 & 63.67 & 80.40 & 73.77 & 75.57 & 70.34 \\
\bottomrule
\end{tabular}%
}

\caption{Cross-source AUC-ROC matrix for KatFishNet (Punctuation + Spacing) baseline under the same cross-source protocol. Dashes indicate same-source entries excluded from evaluation.}
\label{tab:cross_source_baseline}
\end{table*}

\section{Robustness under Decoding Temperature}
\label{sec:appendix_temperature}

We assess whether the proposed features remain effective when the target LLM generates poems at different decoding temperatures. Using GPT-5.2, we generate 189 poems at each of the temperatures $\{0.5, 0.8, 1.0, 1.2\}$ and evaluate the classifier trained on the default GPT-4o source without retraining.

Table~\ref{tab:temperature_stress} reports the per-temperature AUC-ROC values, which range from 95.63 to 98.12. The classifier remains above 95 AUC-ROC across all tested temperatures, indicating that the features capture structural patterns that persist across sampling configurations rather than exploiting temperature-dependent surface artifacts.

\begin{table}[H]
\centering
\setlength{\tabcolsep}{6pt}
\renewcommand{\arraystretch}{1.15}
\begin{tabular}{lc}
\toprule
\textbf{Temperature} & \textbf{AUC-ROC} \\
\midrule
0.5 & 95.63 \\
0.8 & 98.12 \\
1.0 (default) & 96.42 \\
1.2 & 96.39 \\
\bottomrule
\end{tabular}
\caption{Detection AUC-ROC of our method on GPT-5.2 poems generated at varying decoding temperatures. The classifier is trained on the default GPT-4o source without retraining.}
\label{tab:temperature_stress}
\end{table}

\section{Feature Analysis}
\label{sec:appendix_feature_analysis}

\subsection{Raw Volume}
\label{sec:appendix_volume}

\begin{figure}[t!]
    \centering
    \includegraphics[width=0.95\linewidth]{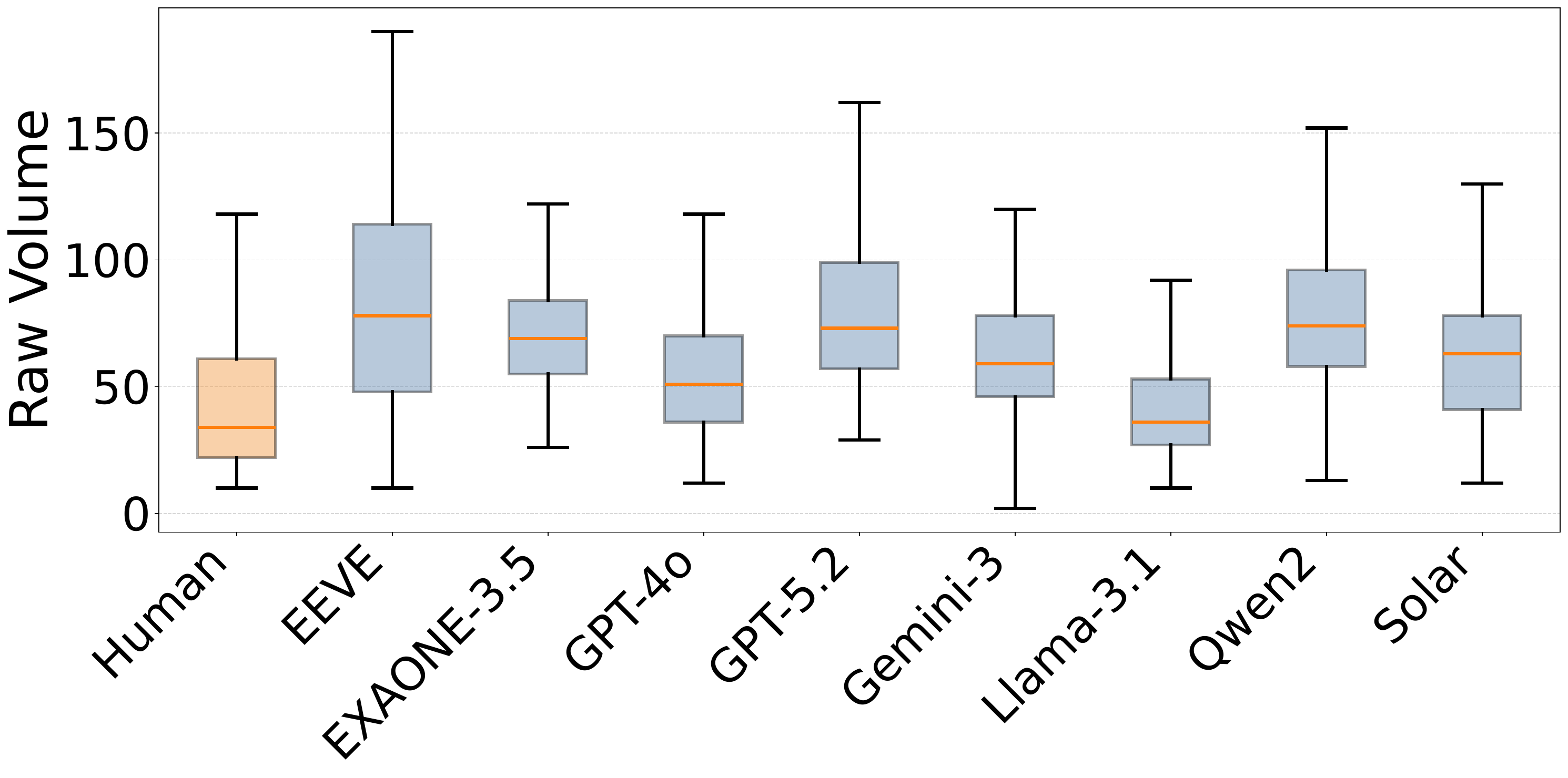}
    \caption{Raw Volume distributions across sources.}
    \label{fig:raw_box}
\end{figure}

Figure~\ref{fig:raw_box} reports Raw Volume, $f_{\text{vol}}$, measured as the whitespace-normalized token count across sources.
It indicates that extremely short outputs are comparatively rare for LLMs, suggesting a weaker preference for compressed forms than in human writing.

\begin{figure}[t!]
    \centering
    \includegraphics[width=0.95\linewidth]{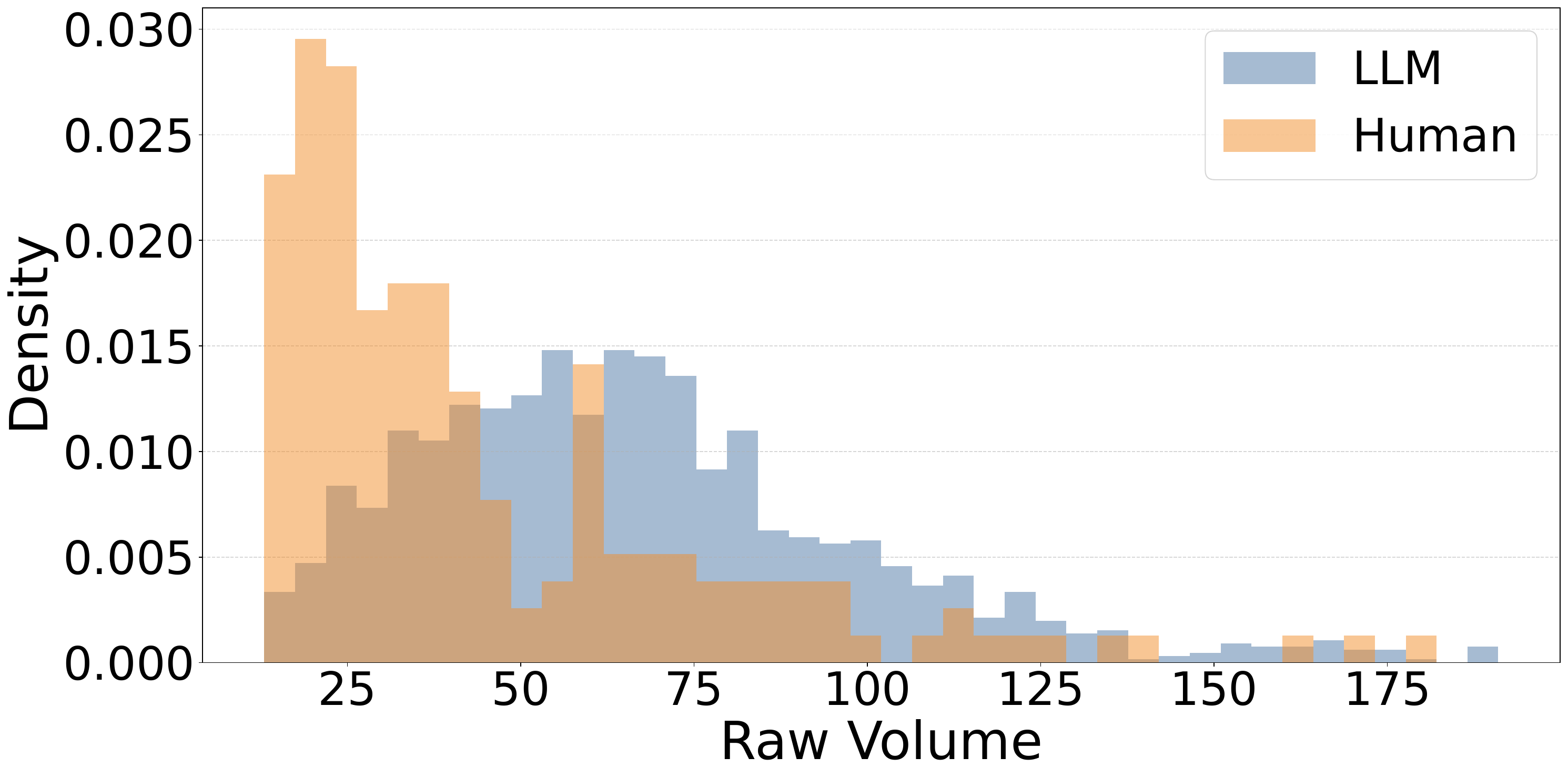}
    \caption{Raw Volume density for human-authored poems and LLM generations.}
    \label{fig:raw_hist}
\end{figure}

Figure~\ref{fig:raw_hist} shows that human-authored poems place substantial probability mass on short outputs, especially $\le30$ tokens, while exhibiting a long-tail over larger volumes.
In contrast, LLM generations are consistently shifted toward longer sequences, yielding higher central tendencies across models.

\subsection{Structure Variation (Ending Type Diversity)}
\label{sec:appendix_structure}

\begin{figure}[t!]
    \centering
    \includegraphics[width=0.95\linewidth]{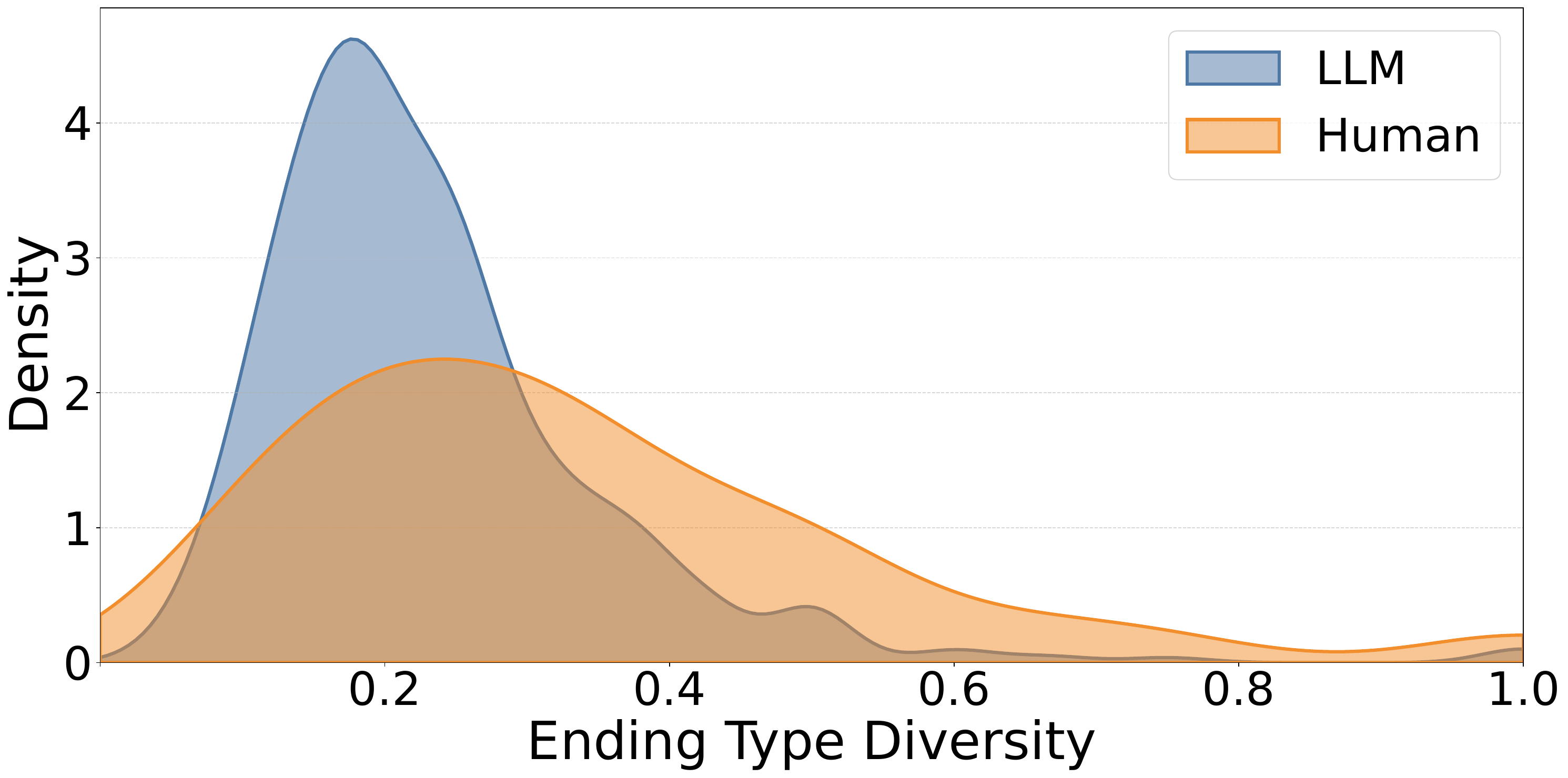}
    \caption{Kernel density of Ending Type Diversity, $f_{\text{div}}$, for human-authored poems and LLM generations.}
    \label{fig:ending_diversity_analysis}
\end{figure}

Figure~\ref{fig:ending_diversity_analysis} summarizes Ending Type Diversity, $f_{\text{div}}$, across human-authored poems and LLM generations.
The density plot shows that LLM outputs concentrate at lower diversity values, whereas human-authored poems exhibit a broader distribution with substantial mass at higher diversity.
This separation indicates that LLM generations more often reuse a limited set of line ending types across a poem, while humans vary sentence-final forms more frequently across lines.

\subsection{Rhythmic Irregularity (Burstiness)}
\label{sec:appendix_rhythm}

\begin{figure}[t!]
    \centering
    \includegraphics[width=0.95\linewidth]{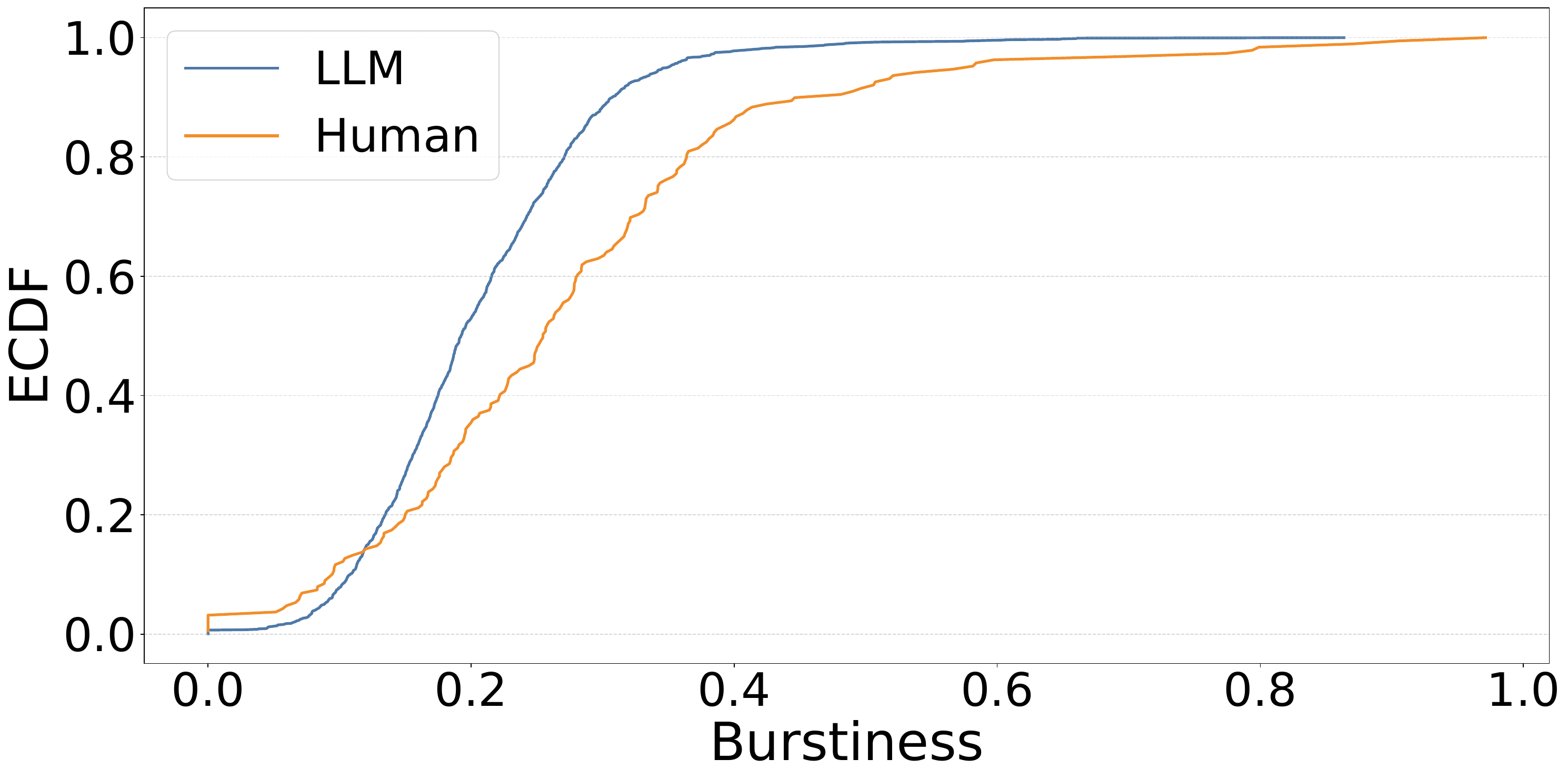}
    \caption{Burstiness ECDF for human-authored poems and LLM generations.}
    \label{fig:burstiness_ecdf}
\end{figure}

Figure~\ref{fig:burstiness_ecdf} illustrates the Burstiness empirical cumulative distribution function (ECDF), $f_{\text{burst}}$, for human-authored poems and LLM generations.
The curve for LLM outputs rises more steeply and approaches one at smaller values, indicating that a significant majority of generated poems exhibit lower burstiness with uniform line-length patterns.
In contrast, human-authored poems demonstrate a slower rise and a longer tail toward higher values, reflecting a distinct tendency for greater rhythmic irregularity and structural variation within the corpus.

\subsection{Normative Adherence}
\label{sec:appendix_normative}

\begin{figure}[t!]
    \centering
    \includegraphics[width=0.95\linewidth]{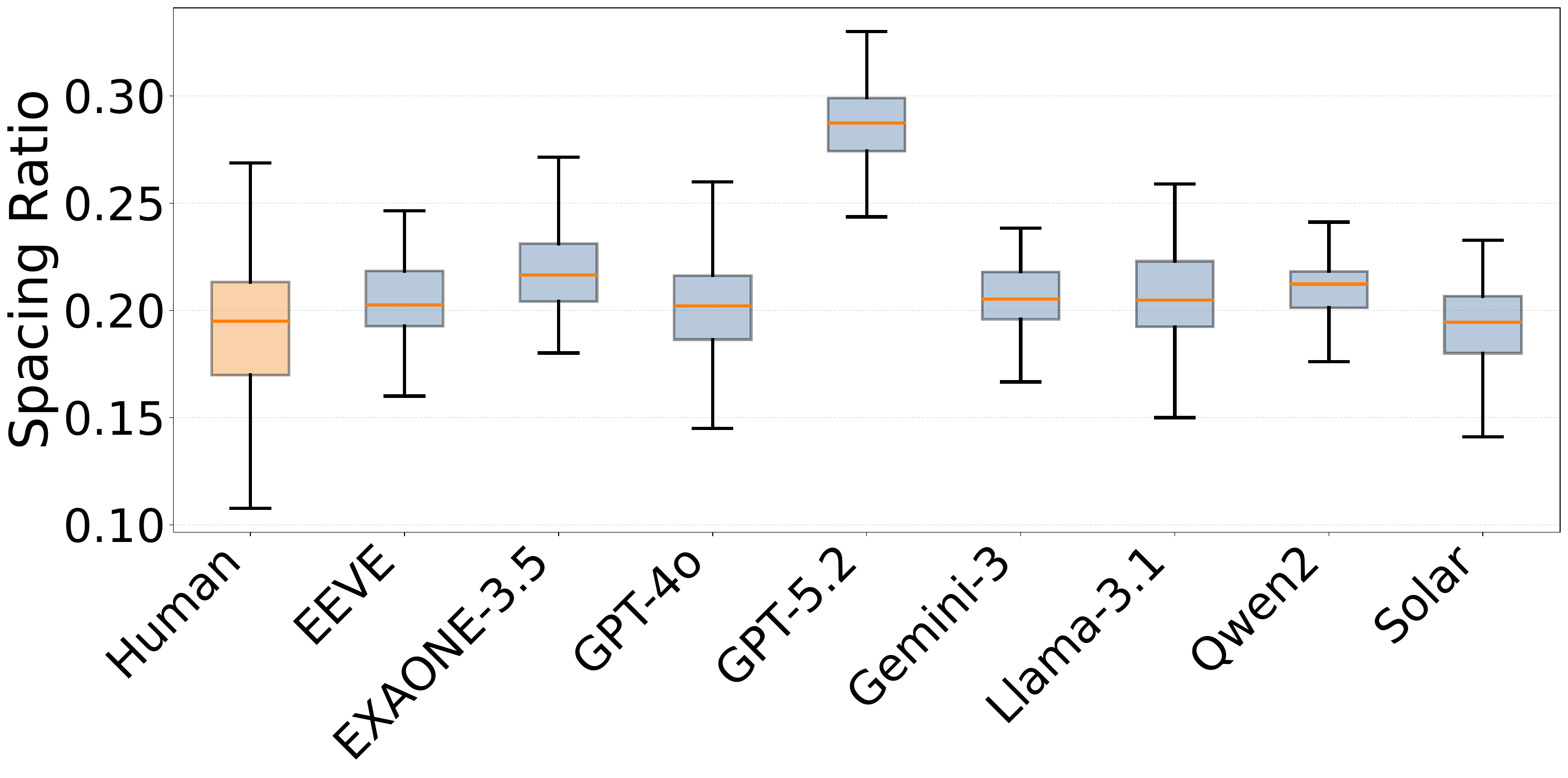}
    \caption{Spacing Ratio distributions across sources.}
    \label{fig:spacing_box}
\end{figure}

Figure~\ref{fig:spacing_box} summarizes the Spacing Ratio, $f_{\text{space}}$, across sources. It shows that the human-authored poems have a lower center and a wider spread, whereas most model outputs cluster more tightly at higher spacing ratios. GPT-5.2 stands out with a markedly higher central tendency than the others, indicating a substantially different spacing pattern.

\begin{figure}[t!]
    \centering
    \includegraphics[width=0.95\linewidth]{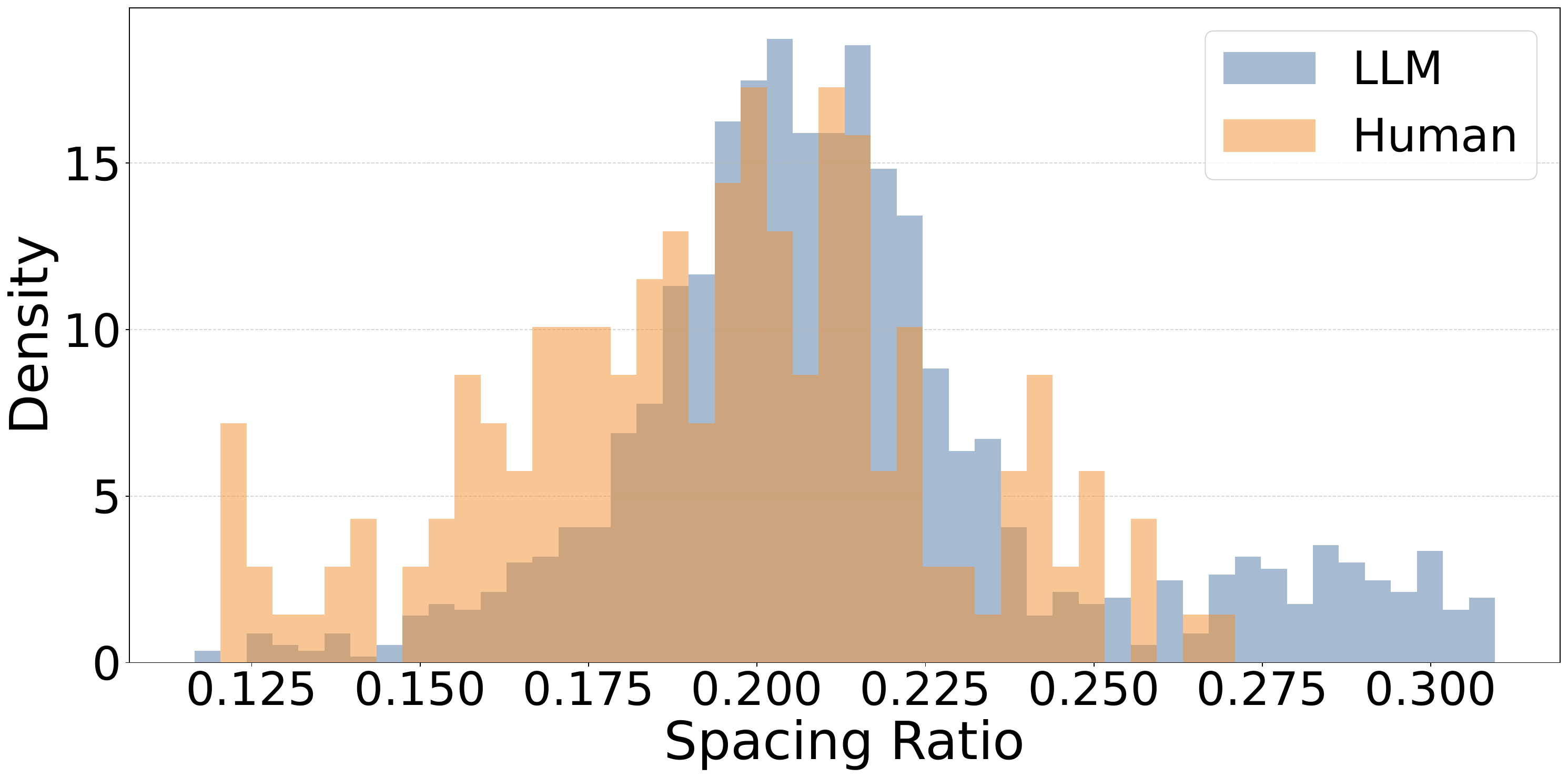}
    \caption{Spacing Ratio density for human-authored poems and LLM generations.}
    \label{fig:spacing_hist}
\end{figure}

Figure~\ref{fig:spacing_hist} shows a right-shifted distribution for model outputs, indicating higher spacing ratios overall.
The upper tail is dominated by model outputs, and values beyond approximately $0.275$ appear only in model-generated texts.

\section{Prompt Robustness and Generation-Side Instruction Ablation}
\label{sec:appendix_prompt_robustness}

Both checks use all 189 GPT-5.2 reference poems and change only the guidance prompt.

\paragraph{Prompt sensitivity.}
Two different LLMs generate paraphrases of the full guidance block while preserving all five instructions. We score the outputs with the detector from Table~\ref{tab:main_logistic} and reuse the GPT-5.2 \textsc{Baseline} and original \textsc{Refined} scores from Table~\ref{tab:main_detection}. Both paraphrases reduce AUC relative to \textsc{Baseline} and bracket the original score. The direction holds across these prompts, although broader prompt robustness remains untested.

\begin{table}[H]
\centering
\small
\begin{tabular}{lc}
\toprule
\textbf{Generation prompt} & \textbf{Detector AUC-ROC} \\
\midrule
\textsc{Baseline}, no guidance & 96.42 \\
\textsc{Refined}, original & 84.28 \\
\textsc{Refined}, paraphrase 1 & 79.46 \\
\textsc{Refined}, paraphrase 2 & 85.00 \\
\bottomrule
\end{tabular}
\caption{Detector AUC-ROC for \textsc{Baseline}, the original \textsc{Refined} prompt, and two instruction-preserving paraphrases on GPT-5.2.}
\label{tab:prompt_paraphrase}
\end{table}

\paragraph{Per-instruction ablation.}
Composite detector AUC cannot isolate one instruction because each omission changes the whole poem. We remove one instruction at a time and regenerate all 189 poems while keeping the remaining guidance fixed. We compare the targeted feature's distance from the human mean with fresh full-guidance outputs from the same run. In this generation run, four instructions move their targeted features toward the human mean, while $f_{\text{burst}}$ is essentially unchanged. Because the features use different scales, we do not rank these distances. The analysis is directional and does not establish statistical significance, additivity, equivalence, or independent causal effects.

\begin{table}[H]
\centering
\small
\setlength{\tabcolsep}{4pt}
\renewcommand{\arraystretch}{1.15}
\begin{tabular}{@{}>{\raggedright\arraybackslash}p{2.7cm} l r r@{}}
\toprule
\textbf{Instruction removed} & \textbf{Target} & \makecell[r]{\textbf{Full}\\\textbf{distance}} & \makecell[r]{\textbf{Removed}\\\textbf{distance}} \\
\midrule
Volume Synchronization & $f_{\text{vol}}$   & 0.49  & 18.93 \\
Rhythmic Variation     & $f_{\text{burst}}$ & 0.018 & 0.017 \\
Connective Reduction   & $f_{\text{conn}}$  & 0.017 & 0.031 \\
Ending Diversity       & $f_{\text{div}}$   & 0.038 & 0.047 \\
Flexible Spacing       & $f_{\text{space}}$ & 0.074 & 0.093 \\
\bottomrule
\end{tabular}
\caption{Target-feature distance from the human mean under full and leave-one-instruction-out guidance on GPT-5.2. Four removals increase the distance, while $f_{\text{burst}}$ remains unchanged.}
\label{tab:instruction_ablation}
\end{table}

\section{Statistical Significance of Human Evaluation}
\label{sec:appendix_friedman}

\begin{figure*}[!t]
    \centering
    \includegraphics[width=0.8\textwidth]{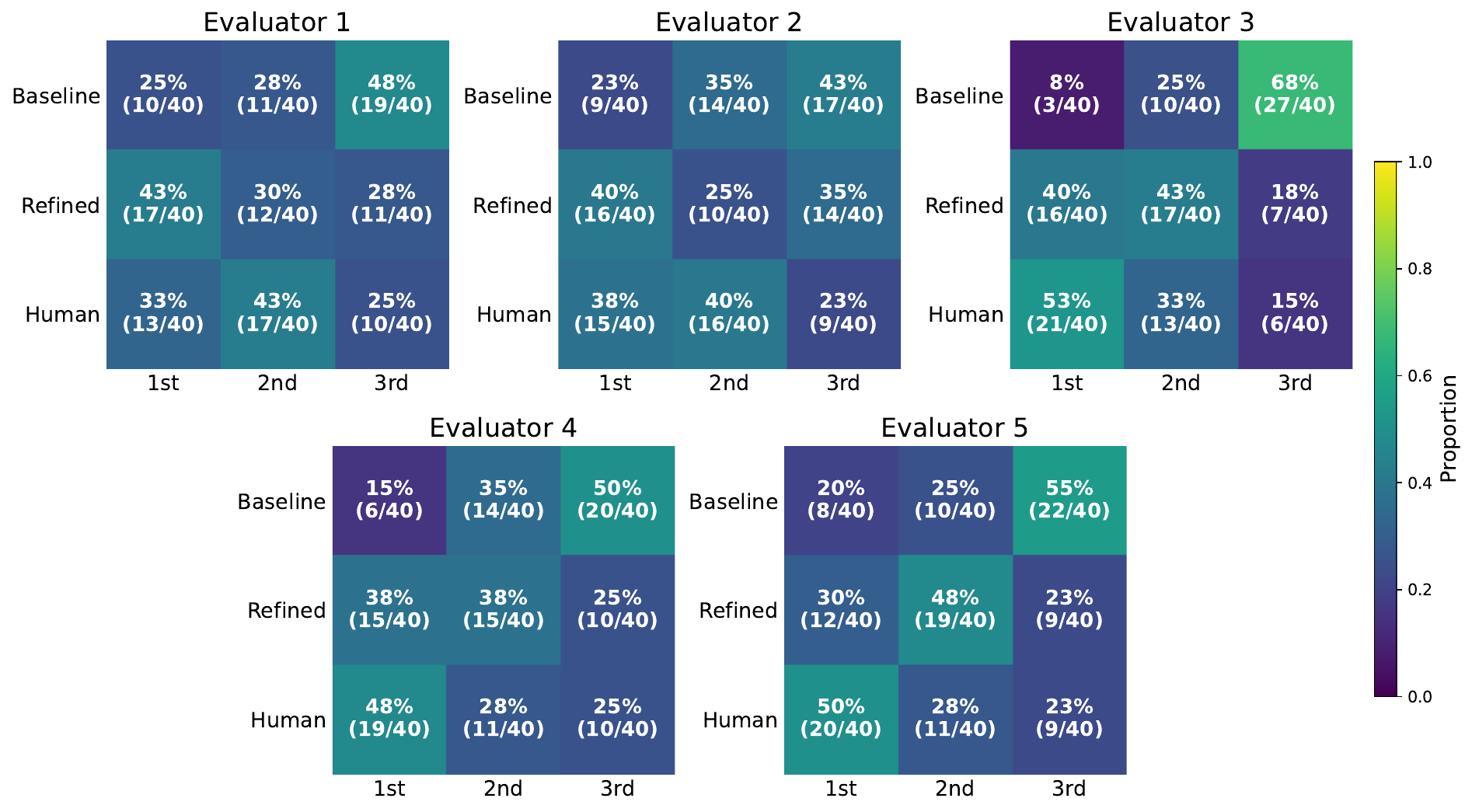}
    \caption{Rank-position distributions over 40 expert-evaluation triplets. Each panel represents one evaluator. Cells show the share of each condition assigned to each rank, with higher ranks indicating greater perceived naturalness.}
    \label{fig:human_eval_rank_heatmap}
\end{figure*}

We use each of the 40 triplets as the paired unit. Within each triplet, we sum the five evaluator scores by condition. We then apply a Friedman test across \textsc{Human}, \textsc{Refined}, and \textsc{Baseline}, followed by pairwise Wilcoxon signed-rank tests with Bonferroni correction.

\begin{itemize}
    \item \textbf{Overall Significance:} Rankings differ across the three conditions at $p < 0.001$.
    \item \textbf{Refined vs.\ Baseline:} \textsc{Refined} receives higher expert naturalness rankings than \textsc{Baseline} at $p < 0.001$.
    \item \textbf{Refined vs.\ Human:} The difference is not significant at $p = 1.0$ after Bonferroni correction. This does not establish statistical equivalence.
    \item \textbf{Baseline vs.\ Human:} \textsc{Human} receives higher scores than \textsc{Baseline} at $p < 0.001$.
\end{itemize}

Figure~\ref{fig:human_eval_rank_heatmap} reports the distribution for each evaluator. \textsc{Baseline} is most often ranked last, while \textsc{Human} and \textsc{Refined} occupy the higher ranks.

These results support only the directional claim that feature guidance narrows the observed gap in perceived naturalness.

\section{Annotator Agreement in Expert Rankings}
\label{app:kripp}

Krippendorff's $\alpha$ is 0.378 for the pooled expert rankings in Section~\ref{sec:main_expert_evaluation}. Low agreement is common in creative-text ranking because evaluators use different criteria and may legitimately disagree~\citep{clark-etal-2021-thats, antoine-etal-2014-weighted,marco-etal-2025-reader}. The rank distributions in Figure~\ref{fig:human_eval_rank_heatmap} nevertheless support the directional \textsc{Refined} over \textsc{Baseline} comparison.

\section[Comparative Analysis of Generation Conditions]{Comparative Analysis of\protect\linebreak\hspace*{\parindent}Generation Conditions}
\label{refined_app:feature_analysis}

\subsection{Volume}
\label{refined_app:volume_dynamics}

\begin{figure}[t!]
    \centering
    \includegraphics[width=0.9\linewidth]{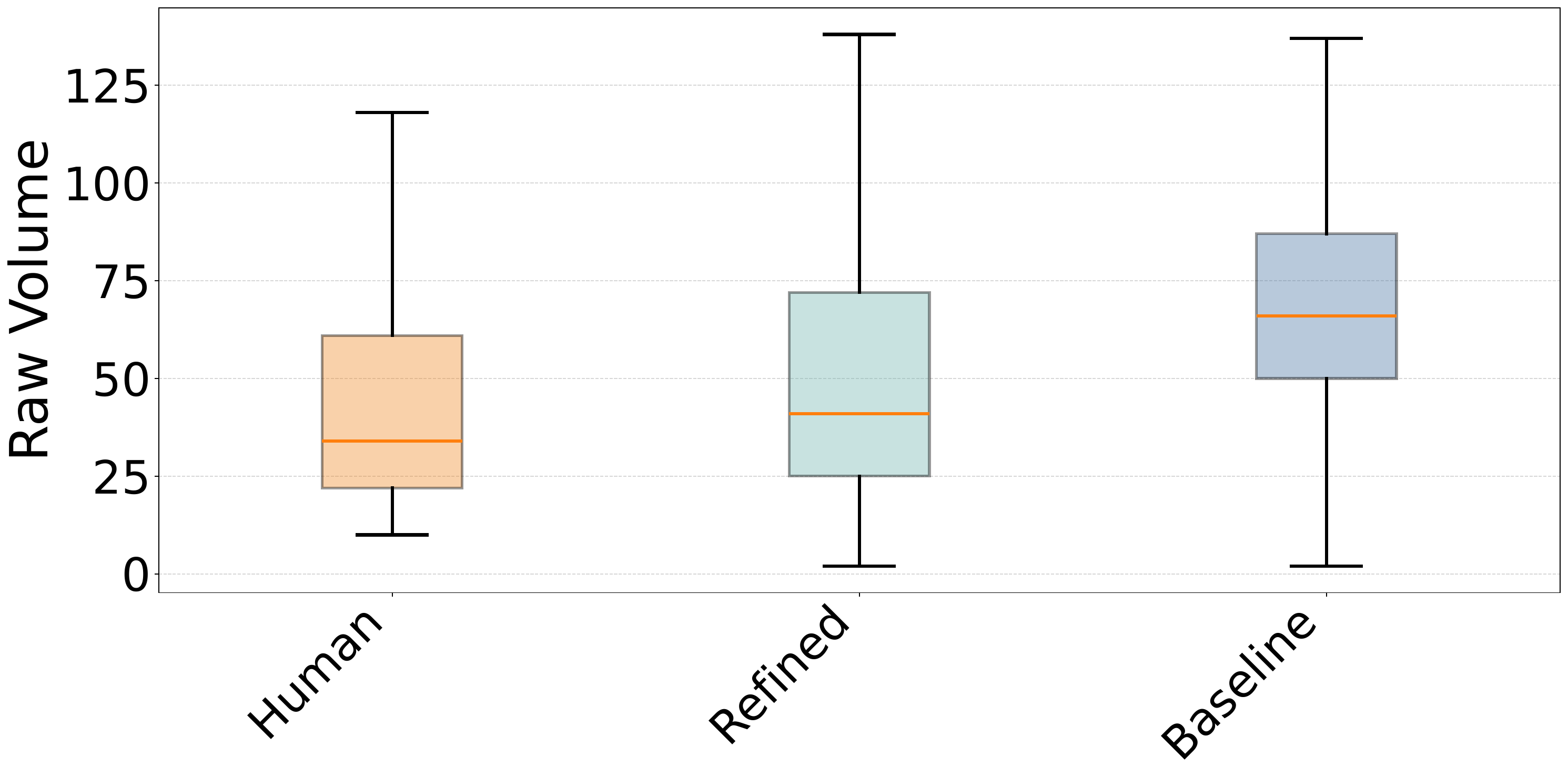}
    \caption{Raw Volume distributions across the \textsc{Human}, \textsc{Baseline}, and \textsc{Refined} conditions.}
    \label{refined_fig:raw_box}
\end{figure}

\begin{figure}[t!]
    \centering
    \includegraphics[width=0.9\linewidth]{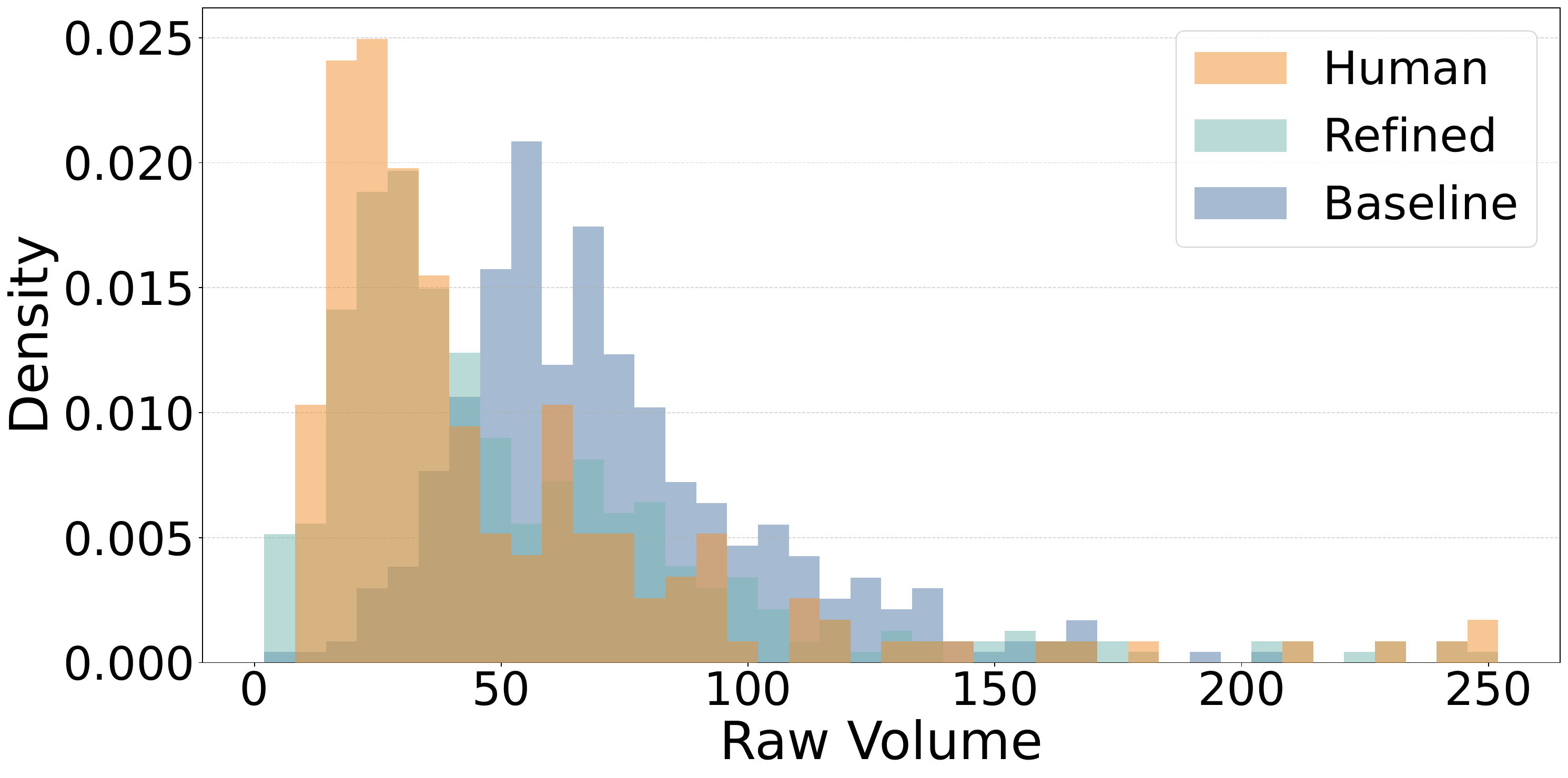}
    \caption{Raw Volume density across the \textsc{Human}, \textsc{Baseline}, and \textsc{Refined} conditions.}
    \label{refined_fig:raw_hist}
\end{figure}

\subsubsection{Raw Volume}
Figure~\ref{refined_fig:raw_box} reports Raw Volume, measured as the whitespace-normalized token count $f_{\text{vol}}$, across poems from the \textsc{Human}, \textsc{Baseline}, and \textsc{Refined} conditions.
It shows that the \textsc{Baseline} distribution is right-shifted with a higher median, whereas \textsc{Refined} moves downward and overlaps more with \textsc{Human}, indicating reduced length inflation.

Figure~\ref{refined_fig:raw_hist} highlights that poems in the \textsc{Human} condition place substantial mass on very short outputs, especially $<25$ tokens, while \textsc{Refined} increases short-to-medium outputs relative to \textsc{Baseline} but still under-represents the shortest region.

\begin{figure}[t!]
    \centering
    \includegraphics[width=0.95\linewidth]{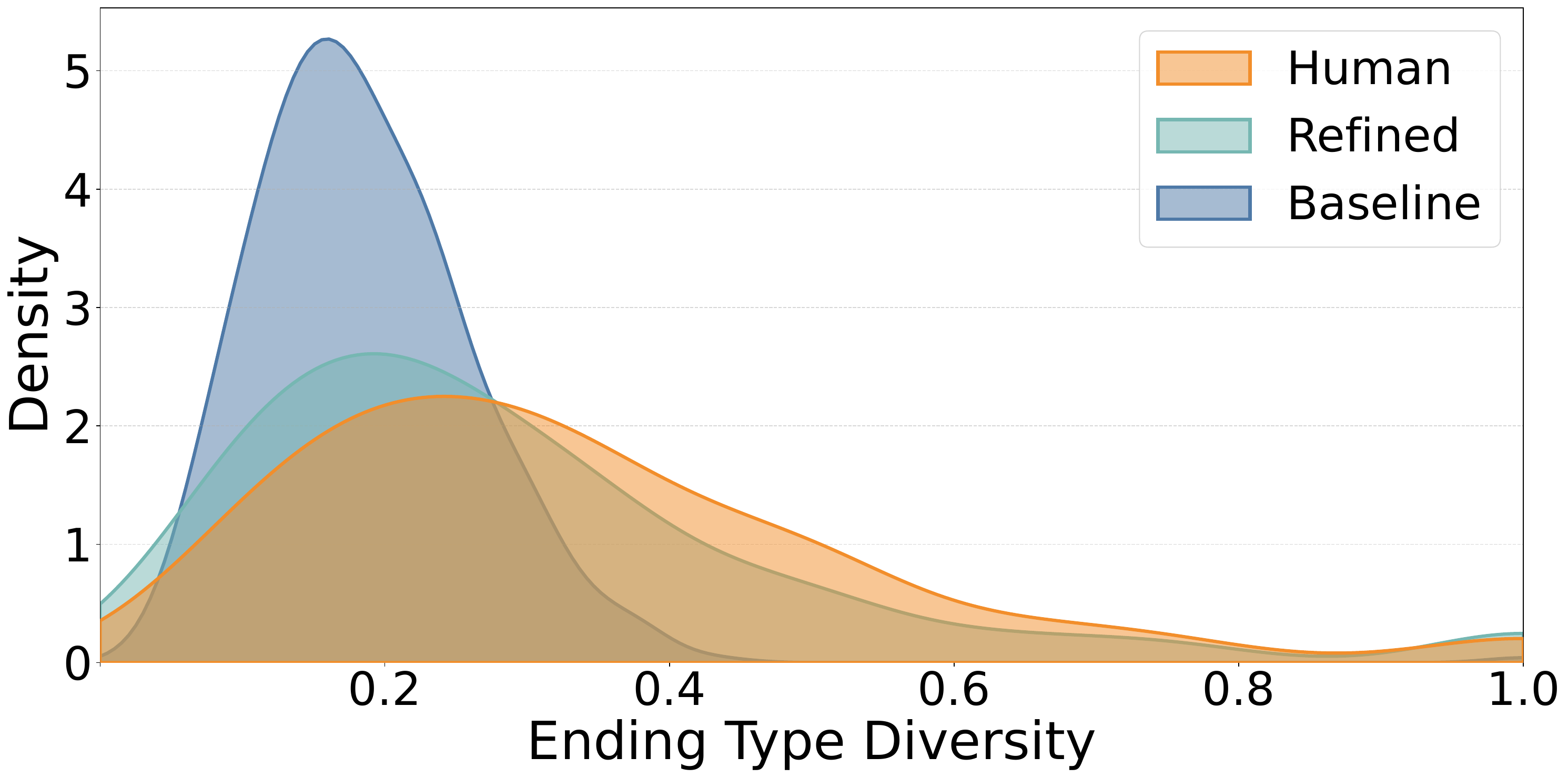}
    \caption{Kernel density of Ending Type Diversity $f_{\text{div}}$ across the \textsc{Human}, \textsc{Baseline}, and \textsc{Refined}.}
    \label{refined_fig:ending_diversity_analysis}
\end{figure}

\subsection{Structure Variation}
\label{refined_sec:appendix_structure}

\subsubsection{Ending Type Diversity}
Figure~\ref{refined_fig:ending_diversity_analysis} summarizes Ending Type Diversity $f_{\text{div}}$ across poems from the \textsc{Human}, \textsc{Baseline}, and \textsc{Refined} conditions.
The density plot shows that \textsc{Baseline} concentrates at low diversity values with a sharp peak, indicating frequent reuse of a limited set of line-ending types within a poem.
In contrast, \textsc{Refined} shifts mass toward higher diversity and broadens the distribution, suggesting that feature guidance encourages more varied sentence-final forms across lines.
Nevertheless, the \textsc{Human} condition exhibits the broadest spread with a heavier right tail, implying that human-authored poems still realize greater ending-type variability than guided generations.

\subsection{Normative Adherence}
\label{refined_sec:appendix_normative}

\begin{figure}[ht]
    \centering
    \includegraphics[width=0.95\linewidth]{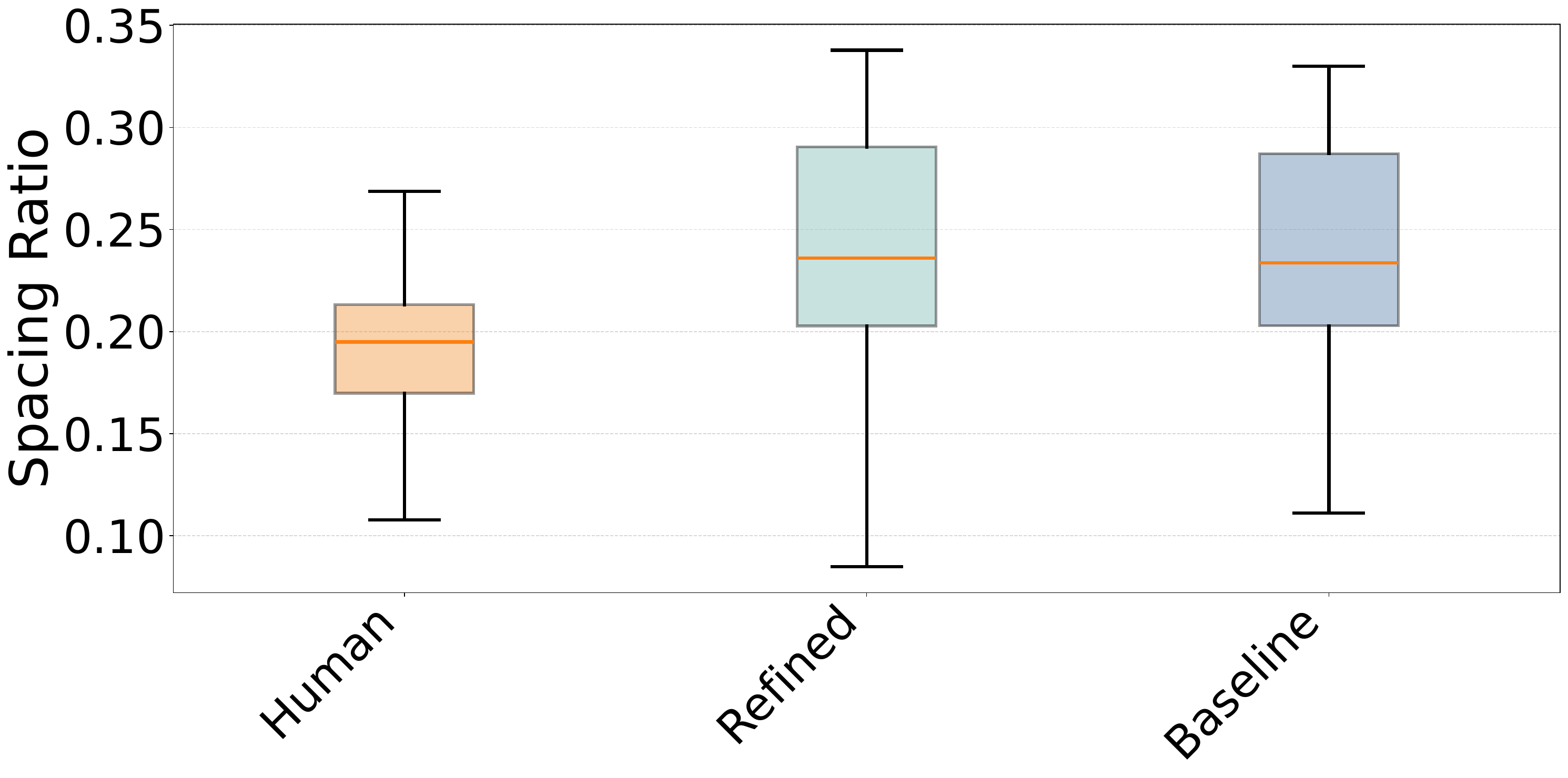}
    \caption{Spacing Ratio distributions across the \textsc{Human}, \textsc{Baseline}, and \textsc{Refined} conditions.}
    \label{refined_fig:spacing_box}
\end{figure}

Figure~\ref{refined_fig:spacing_box} summarizes the Spacing Ratio, $f_{\text{space}}$, across poems from the \textsc{Human}, \textsc{Baseline}, and \textsc{Refined} conditions.
Poems in the \textsc{Human} condition generally exhibit a lower center whereas \textsc{Baseline} remains around higher values.
In contrast, \textsc{Refined} demonstrates a marked increase in variance, as observed in both the boxplot and the histogram.

\begin{figure}[ht]
    \centering
    \includegraphics[width=0.95\linewidth]{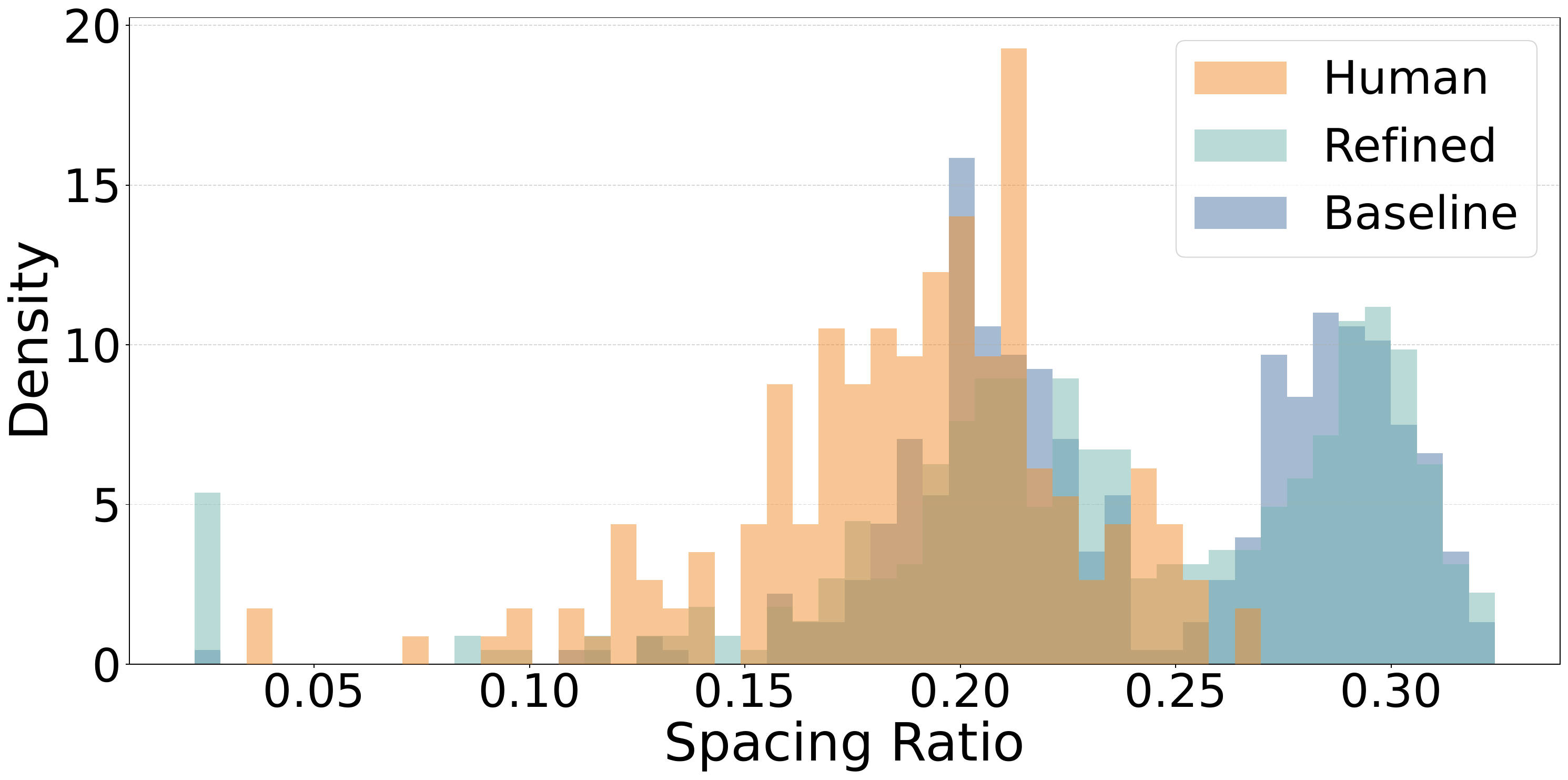}
    \caption{Spacing Ratio density across the \textsc{Human}, \textsc{Baseline}, and \textsc{Refined} conditions.}
    \label{refined_fig:spacing_hist}
\end{figure}

Figure~\ref{refined_fig:spacing_hist} reveals that the distribution of \textsc{Refined} spreads out significantly, often producing spacing ratios even lower than those of the \textsc{Human} condition, yet it still maintains substantial overlap with \textsc{Baseline} in the higher-ratio region.
However, the improvement remains modest compared to other features, likely due to the models' inherent adherence to standard orthography~\citep{park-etal-2025-katfishnet,wu-etal-2025-wrote}.

\section{Implementation Details}
\label{sec:appendix_impl_mgtbench}

This appendix details the detector benchmark used in Section~\ref{sec:detectors}. When reproducing baselines, we use the implementations provided by MGTBench-2.0~\citep{Liu_2025}.

\begin{itemize}
    \item \textsc{Log-Likelihood} \citep{solaiman2019releasestrategiessocialimpacts}:
    assigns a score to a passage by averaging token log-probability under the \texttt{EleutherAI/gpt-neo-2.7B}~\citep{gao2020pile}\footnote{\label{fn:neo}\url{https://huggingface.co/EleutherAI/gpt-neo-2.7B}} as the scoring model.

    \item \textsc{Rank} \citep{gehrmann-etal-2019-gltr}:
    computes the average rank of the observed next token under the \texttt{EleutherAI/gpt-neo-2.7B}\footref{fn:neo} for scoring.

    \item \textsc{Entropy} \citep{gehrmann-etal-2019-gltr}:
    measures predictive uncertainty by averaging token-level entropy of the next-token distribution from \texttt{EleutherAI/gpt-neo-2.7B}\footref{fn:neo}.

    \item \textsc{LRR} (Log-Likelihood Log-Rank Ratio) \citep{su-etal-2023-detectllm}:
    defines a perturbation-free statistic derived from token rank information, implemented as a log-rank ratio score based on \texttt{EleutherAI/gpt-neo-2.7B}\footref{fn:neo}.

    \item \textsc{NPR} (Normalized Perturbed log-Rank) \citep{su-etal-2023-detectllm}:
    quantifies how token-rank signals change under small perturbations, utilizing \texttt{google/mt5-xl}~\citep{xue2021mt5}\footnote{\label{fn:mt5}\url{https://huggingface.co/google/mt5-xl}} as the masking model alongside \texttt{EleutherAI/gpt-neo-2.7B}\footref{fn:neo} for scoring.

    \item \textsc{Binoculars} \citep{pmlr-v235-hans24a}:
    scores texts by comparing surprisal values between two related language models, configured with \texttt{Qwen/Qwen2.5-7B}~\citep{yang2024qwen25}\footnote{\label{fn:qwen}\url{https://huggingface.co/Qwen/Qwen2.5-7B}} as the observer and \texttt{Qwen/Qwen2.5-7B-Instruct}\footnote{\label{fn:qweninst}\url{https://huggingface.co/Qwen/Qwen2.5-7B-Instruct}} as the performer.

    \item \textsc{Fast-DetectGPT} \citep{bao2024fastdetectgpt}:
    provides an efficient curvature-based detector in the DetectGPT family, instantiated with \texttt{EleutherAI/gpt-neo-2.7B}\footref{fn:neo} as the scoring model and \texttt{EleutherAI/gpt-j-6B}~\citep{gao2020pile}\footnote{\label{fn:gptj}\url{https://huggingface.co/EleutherAI/gpt-j-6b}} as the reference model.
\end{itemize}

\onecolumn

\section{Glossary of Terms and Abbreviations}
\label{sec:appendix_glossary}

`KatFishNet' is the closest prior work on Korean LLM-text detection and the strongest baseline in our comparison, built on punctuation and spacing cues; it also provides the human and four-LLM dataset we build on. `OOD detection' means a zero-shot out-of-distribution protocol in which a classifier is trained on human poems and one generator, then evaluated on unseen generators.

`MMD' (Maximum Mean Discrepancy) is a kernel-based distance between two sample distributions. `RBF kernel' is a radial basis function kernel used for the MMD tests. `Borda Count' is a rank aggregation assigning 2, 1, and 0 points to first, second, and third place. `Kkma' is a Korean morphological analyzer and POS tagger. `AUC-ROC' is the area under the receiver operating characteristic curve. `DeLong test' is a statistical test for comparing two correlated AUC-ROC values.

\par\vspace{1em}


\section{Prompt}
\label{sec:prompt}
Figure~\ref{fig:prompt_constraints} presents the prompt used for poetry generation. 
The gray text follows the KatFishNet~\citep{park-etal-2025-katfishnet} baseline, conditioning on an age-group persona and a given poem while restricting outputs to Korean only. 
The blue text adds feature-based guidance that encourages length alignment, line-length variation, reduced connective endings, diverse line-final forms, and flexible spacing.

\definecolor{kfbase}{RGB}{90,90,90}      
\definecolor{ouradd}{RGB}{20,70,150}   
\begin{figure*}[h]
\small
\centering
\fbox{%
  \begin{minipage}{0.9\textwidth}
    {\color{kfbase}
    You are a poet in your \{\textit{age\_group}\}.\\
    Read the given poem and understand its content. Then, write a new poem in your style that suits your age group.\\ 
    Write only in Korean. Output the new poem only.\\ 
    }%
    \\
    {\color{ouradd}
    You must follow the structural constraints below.\\
    \\
    1. Volume Synchronization\\
    - Keep the total length, especially the number of lines, close to that of the given poem.\\ 
    \\
    2. Rhythmic Variation\\
    - To introduce rhythmic irregularity, do not make all lines similar in length.\\
    \\
    3. Connective Reduction\\
    - Do not keep linking clauses in a long sequence using endings like ``-하고'' (\textit{-hago}, ``and'') or ``-하며'' (\textit{-hamyeo}, ``while'').\\
    - Reduce conjunctions. Instead, cut sentences more decisively or connect meaning through line breaks.\\
    \\
    4. Ending Diversity\\
    - Avoid ending every line in the same style.\\
    - Use diverse closing forms, including nominal endings and other variants.\\
    \\
    5. Flexible spacing\\
    - Do not aim for perfectly standard spacing.\\
    - It is acceptable to mimic occasional spacing quirks or small deviations that appear in human writing.\\
    }%
    \\
    {\color{kfbase}
    Write only in Korean. Output the new poem only.\\
    \\
    Given Poem:\\ 
    \{\textit{poetry}\}
    }%
  \end{minipage}%
}
\caption{Prompt used for poetry generation} 
\label{fig:prompt_constraints}
\end{figure*}
\clearpage
\onecolumn
\section{Human Evaluation Questionnaire}
\label{sec:human_eval_questionnaire}

Figures~\ref{fig:questionnaire_example_korean} and \ref{fig:questionnaire_example_english} present the interface used for the blind expert evaluation.
For each question, evaluators were shown a triplet of poems consisting of the human-authored original, the \textsc{Baseline} generation, and the \textsc{Refined} generation.
To prevent bias, the presentation order within each triplet was fully randomized.
Evaluators were instructed to read all three poems and rank them based on their stylistic naturalness and resemblance to human-written poetry.

\par\vspace{1em}

\begin{figure*}[h]
    \centering
    \fbox{%
    \begin{minipage}{0.9\textwidth}
        \vspace{0.5em}
        \textbf{Q18. 다음 시들 중 `가장 사람이 쓴 것처럼 자연스러운 시' 순서대로 순위를 매겨주세요.}
        \par\vspace{0.5em}
        \hrule
        \par\vspace{1em}
        
        \begin{minipage}[t]{0.31\textwidth}
            \centering\textbf{(A)}
            \par\vspace{0.5em}
            \small
            \raggedright
            좋아한다는 말은 어리숙해 보여 참았고,\\
            보고 싶다는 말은 나만 애타는 것 같아 말았다.\\
            그립다는 말은 서랍 속 빛 바랜 사진 같았고,\\
            사랑한다는 말은 너무 뜨거워 도저히 뱉을 수 없다.\\
            더이상 형용할 수 없는 사람의, 사랑의 말.
        \end{minipage}%
        \hfill\vrule\hfill 
        \begin{minipage}[t]{0.31\textwidth}
            \centering\textbf{(B)}
            \par\vspace{0.5em}
            \small
            \raggedright
            좋아한다는 말은\\
            급식 줄에서 내 순서 기다리다\\
            괜히 목에 걸려 삼켰고,\\
            \vspace{0.3em}
            ...\\
            \vspace{0.3em}
            그래서 나는\\
            네 이름을\\
            알람처럼 매일 울리게 해 두고,\\
            \vspace{0.3em}
            아무 말 대신\\
            오늘도 너한테 가는 길을\\
            조용히 저장한다.
        \end{minipage}%
        \hfill\vrule\hfill 
        \begin{minipage}[t]{0.31\textwidth}
            \centering\textbf{(C)}
            \par\vspace{0.5em}
            \small
            \raggedright
            좋아한단 말은 학생증 사진처럼 촌스러워 숨겼고\\
            보고 싶단 말은 내 채팅창만 새로고침하는 것 같아 지웠다\\
            그립단 말은 공책 뒤에 끼워둔 영화표\\
            구겨진 채로 남아\\
            사랑한단 말은 손바닥에 쥔 불씨\\
            입술에 닿기 전부터 아프다\\
            더는 이름 붙일수 없는 너의, 내 말의 끝.
        \end{minipage}
        
        \par\vspace{1.5em}
        \hrule
        \par\vspace{1em}
        
        \centering
        \textbf{순위 선택:} \quad
        1위: \underline{\hspace{2cm}} \quad
        2위: \underline{\hspace{2cm}} \quad
        3위: \underline{\hspace{2cm}}
        \vspace{0.5em}
    \end{minipage}%
    }
    \caption{Korean questionnaire used to rank randomized \textsc{Human}, \textsc{Baseline}, and \textsc{Refined} triplets by naturalness.}
    \label{fig:questionnaire_example_korean}
\end{figure*}

\par\vspace{1em}

\begin{figure*}[h]
    \centering
    \fbox{%
    \begin{minipage}{0.9\textwidth}
        \vspace{0.5em}
        \textbf{Q18. Please read the poems below and rank them in order of naturalness as human-authored poetry.}
        \par\vspace{0.5em}
        \hrule
        \par\vspace{1em}
        
        \begin{minipage}[t]{0.31\textwidth}
            \centering\textbf{(A)}
            \par\vspace{0.5em}
            \small
            \raggedright
            I held back saying ``I like you'' because it looked clumsy,\\
            I stopped saying ``I miss you'' because it felt like only I was anxious.\\
            ``I long for you'' was like a faded photo in a drawer,\\
            ``I love you'' is too hot, I can't spit it out.\\
            Words of love, of a person I can no longer describe.
        \end{minipage}%
        \hfill\vrule\hfill 
        \begin{minipage}[t]{0.31\textwidth}
            \centering\textbf{(B)}
            \par\vspace{0.5em}
            \small
            \raggedright
            The words ``I like you''\\
            waiting for my turn in the school lunch line\\
            caught in my throat, so I swallowed them,\\
            \vspace{0.5em}
            ...\\
            \vspace{0.5em}
            So I\\
            make your name\\
            ring every day like an alarm,\\
            \vspace{0.5em}
            and instead of saying anything\\
            today, too, I quietly save\\
            the path to you.
        \end{minipage}%
        \hfill\vrule\hfill 
        \begin{minipage}[t]{0.31\textwidth}
            \centering\textbf{(C)}
            \par\vspace{0.5em}
            \small
            \raggedright
            I hid ``I like you'' because it was tacky like a student ID photo\\
            I deleted ``I miss you'' because it looked like I was just refreshing my chat window\\
            ``I long for you'' is a movie ticket stuck behind a notebook\\
            remaining crumpled\\
            ``I love you'' is a spark held in the palm\\
            hurting before it even touches the lips\\
            The end of my words, of you who I can no longer name.
        \end{minipage}
        
        \par\vspace{1.5em}
        \hrule
        \par\vspace{1em}
        
        \centering
        \textbf{Rank Selection:} \quad
        1st: \underline{\hspace{2cm}} \quad
        2nd: \underline{\hspace{2cm}} \quad
        3rd: \underline{\hspace{2cm}}
        \vspace{0.5em}
    \end{minipage}%
    }
    \caption{English translation of Figure~\ref{fig:questionnaire_example_korean}, preserving the poems' line structure.}
    \label{fig:questionnaire_example_english}
\end{figure*}

\end{document}